\documentclass[10pt,twocolumn,letterpaper]{article}

\usepackage{iccv}
\usepackage{times}
\usepackage{epsfig}
\usepackage{graphicx}
\usepackage{amsmath}
\usepackage{amssymb}

\usepackage{booktabs}
\usepackage{capt-of}
\usepackage{tabularx}
\newcolumntype{C}{>{\centering\arraybackslash}X}
\newcolumntype{L}{>{\raggedright\arraybackslash}X}
\newcolumntype{R}{>{\raggedleft\arraybackslash}X}
\makeatletter
\let\MYcaption\@makecaption
\makeatother
\usepackage[subrefformat=parens]{subcaption}
\makeatletter
\let\@makecaption\MYcaption
\makeatother
\makeatletter
\newcommand\footnoteref[1]{\protected@xdef\@thefnmark{\ref{#1}}\@footnotemark}
\makeatother

\usepackage[pagebackref=true,breaklinks=true,colorlinks,bookmarks=false]{hyperref}
\usepackage{mmstyles}
\usepackage{pifont}
\usepackage[table,dvipsnames]{xcolor}
\usepackage[absolute]{textpos}
\usepackage{appendix}
\usepackage{siunitx}
\usepackage[capitalize]{cleveref}
\usepackage[utf8]{inputenc} 
\usepackage[T1]{fontenc}    
\usepackage{url}            
\usepackage{amsfonts}       
\usepackage{nicefrac}       
\usepackage{microtype}      
\usepackage{diagbox}
\usepackage{wrapfig,lipsum}
\usepackage{multirow}
\usepackage{multicol}
\usepackage{listings}
\usepackage{color}
\usepackage{mathtools}
\usepackage{paralist} 
\usepackage{colortbl}
\usepackage{makecell}
\usepackage[outline]{contour}
\usepackage{soul}
\usepackage{bbm}
\usepackage[accsupp]{axessibility}  
\usepackage[percent]{overpic}
\usepackage{pdfpages}
\usepackage{tikz}

\usepackage[breaklinks=true,bookmarks=false]{hyperref}


\iccvfinalcopy 

\def\iccvPaperID{2330} 

 \newcommand{\supp}[1]{{{#1}}}

\ificcvfinal\fi

\begin{document}
\definecolor{yellow}{rgb}{1,1, 0.6}
\definecolor{lightyellow}{rgb}{1,1, 0.8}
\definecolor{orange}{rgb}{1, 0.8, 0.6}
\definecolor{red}{rgb}{1, 0.6, 0.6}

\definecolor{wincolor}{rgb}{0.85, 0.0, 0.0}

\definecolor{darkyellow}{rgb}{0.8, 0.8, 0.5}
\definecolor{darkred}{rgb}{0.7, 0.3, 0.3}
\definecolor{darkgreen}{rgb}{0.3, 0.7, 0.3}
\definecolor{blue}{rgb}{0, 0, 1.0}
\definecolor{green}{rgb}{0, 1.0, 0}
\definecolor{pink}{rgb}{1, 0.4, 0.7}

\newcommand{\barron}[1]{{\color{blue} barron: #1}}
\newcommand{\ricardo}[1]{{\color{darkgreen} ricardo: #1}}
\newcommand{\todo}[1]{{\color{pink} TODO: #1}}

\newcommand{\mbf}[1]{{\mathbf{#1}}}

\let\originalleft\left
\let\originalright\right
\renewcommand{\left}{\mathopen{}\mathclose\bgroup\originalleft}
\renewcommand{\right}{\aftergroup\egroup\originalright}

\newcommand{\norm}[1]{\left\lVert#1\right\rVert}

\newcommand{\expo}[1]{\exp\left(#1\right)}

\newcommand{\modeltheta}{\mathrm{\Theta}}
\newcommand{\absrp}{\sigma}

\newcommand{\numsamples}{N}
\newcommand{\numsamplescoarse}{N_c}
\newcommand{\numsamplesfine}{N_f}
\newcommand{\timenear}{t_n}
\newcommand{\timefar}{t_f}
\newcommand{\deltatime}{\delta}

\newcommand{\posxy}{xy}
\newcommand{\posxyz}{xyz}
\newcommand{\angletheta}{\theta}
\newcommand{\anglephi}{\phi}
\newcommand{\posall}{\posxyz\angletheta\anglephi}

\newcommand{\numfrequencies}{L}

\newcommand{\Ltrain}{\mathcal{L}}
\newcommand{\raybatch}{\mathcal{R}}
\newcommand{\Ccoarse}{\hat{C}_c(\ray)}
\newcommand{\Cfine}{\hat{C}_f(\ray)}
\newcommand{\Ctrue}{C(\ray)}
\newcommand{\pweight}{w}
\newcommand{\normpweight}{\hat{w}}

\newcommand{\scenename}[1]{\textit{#1}}

\newcommand{\shortpara}[1]{{\bf #1}}

\newcommand{\longmodelname}{\emph{multum in parvo} NeRF\xspace}
\newcommand{\shortmodelname}{Mip-NeRF\xspace}
\newcommand{\shortmodelnamelower}{mip-NeRF\xspace}

%
%

%
%
%
%
\newcommand{\cone}{\mathbf{\mathcal{C}}}
\newcommand{\sphere}{\mathbf{\mathcal{S}}}
\newcommand{\aabb}{\mathbf{\mathcal{B}}}
\newcommand{\rayorigin}{\mathbf{o}}
\newcommand{\raydir}{\mathbf{d}}
\newcommand{\vieworigin}{\mathbf{o}}
\newcommand{\viewdir}{\mathbf{d}}
\newcommand{\pixcenter}{\mathbf{p_o}}
\newcommand{\feat}{\mathbf{f}}
\newcommand{\planeXY}{\text{plane}_{XY}}
\newcommand{\planeXZ}{\text{plane}_{XZ}}
\newcommand{\planeYZ}{\text{plane}_{YZ}}
\newcommand{\disc}{\mathbf{\mathcal{D}}}

\newcommand{\modelweights}{\Theta}
\newcommand{\mipmap}{\mathcal{M}}

\newcommand{\density}{\tau}
\newcommand{\col}{{c}}
\newcommand{\Col}{\mathbf{C}}

\newcommand{\position}{\mathbf{x}}
\newcommand{\posenc}{\gamma(\mathbf{x})}
\newcommand{\radius}{\mathbf{r}}
\newcommand{\baseradius}{\dot r}
\newcommand{\featradius}{\ddot r}
\newcommand{\mlp}{\operatorname{MLP}}
\newcommand{\trimip}{\operatorname{Tri-Mip}}

\newcommand{\zval}{t}
\newcommand{\zvec}{\mathbf{\zval}}
\newcommand{\ray}{\mathbf{r}}
\newcommand{\transpose}{{\operatorname{T}}}
\newcommand{\myparagraph}[1]{\vspace{0.5em} \noindent {\bf #1}\,\,\,}
\newcommand{\thousand}{K}
\newcommand{\million}{M}
%
%

\title{
Tri-MipRF: Tri-Mip Representation for Efficient Anti-Aliasing \\ Neural Radiance Fields
}

\author{
Wenbo Hu$^{1}$
\quad 
Yuling Wang$^{1,2}$ 
\quad 
Lin Ma$^{1}$
\quad 
Bangbang Yang$^{1}$ 
\\
Lin Gao$^{3}$
\quad 
Xiao Liu$^{1}$
\quad 
Yuewen Ma$^{1\dag}$ \\
	$^{1}$PICO, ByteDance, Beijing \quad $^{2}$Tsinghua University \quad $^{3}$Institute of Computing Technology, CAS\\
}

\maketitle  
\renewcommand{\thefootnote}{\fnsymbol{footnote}}
 \footnotetext[2]{Corresponding author.}

\ificcvfinal\thispagestyle{empty}\fi


\begin{abstract}
Despite the tremendous progress in neural radiance fields (NeRF), we still face a dilemma of the trade-off between quality and efficiency, e.g., MipNeRF~\cite{barron2021mip} presents fine-detailed and anti-aliased renderings but takes days for training, while Instant-ngp~\cite{muller2022instant} can accomplish the reconstruction in a few minutes but suffers from blurring or aliasing when rendering at various distances or resolutions due to ignoring the sampling area.
To this end, we propose a novel $\trimip$ encoding (à la ``mipmap'') that enables both instant reconstruction and anti-aliased high-fidelity rendering for neural radiance fields. 
The key is to factorize the pre-filtered 3D feature spaces in three orthogonal mipmaps.
In this way, we can efficiently perform 3D area sampling by taking advantage of 2D pre-filtered feature maps, which significantly elevates the rendering quality without sacrificing efficiency.
To cope with the novel $\trimip$ representation, we propose a cone-casting rendering technique to efficiently sample anti-aliased 3D features with the $\trimip$ encoding considering both pixel imaging and observing distance.
%
%
Extensive experiments on both synthetic and real-world datasets demonstrate our method achieves state-of-the-art rendering quality and reconstruction speed while maintaining a compact representation that reduces $25\%$ model size compared against Instant-ngp.
Code is available at the project webpage: \url{https://wbhu.github.io/projects/Tri-MipRF}
\end{abstract}


\begin{figure}[!t] 
	\centering
	\includegraphics[width=0.93\linewidth]{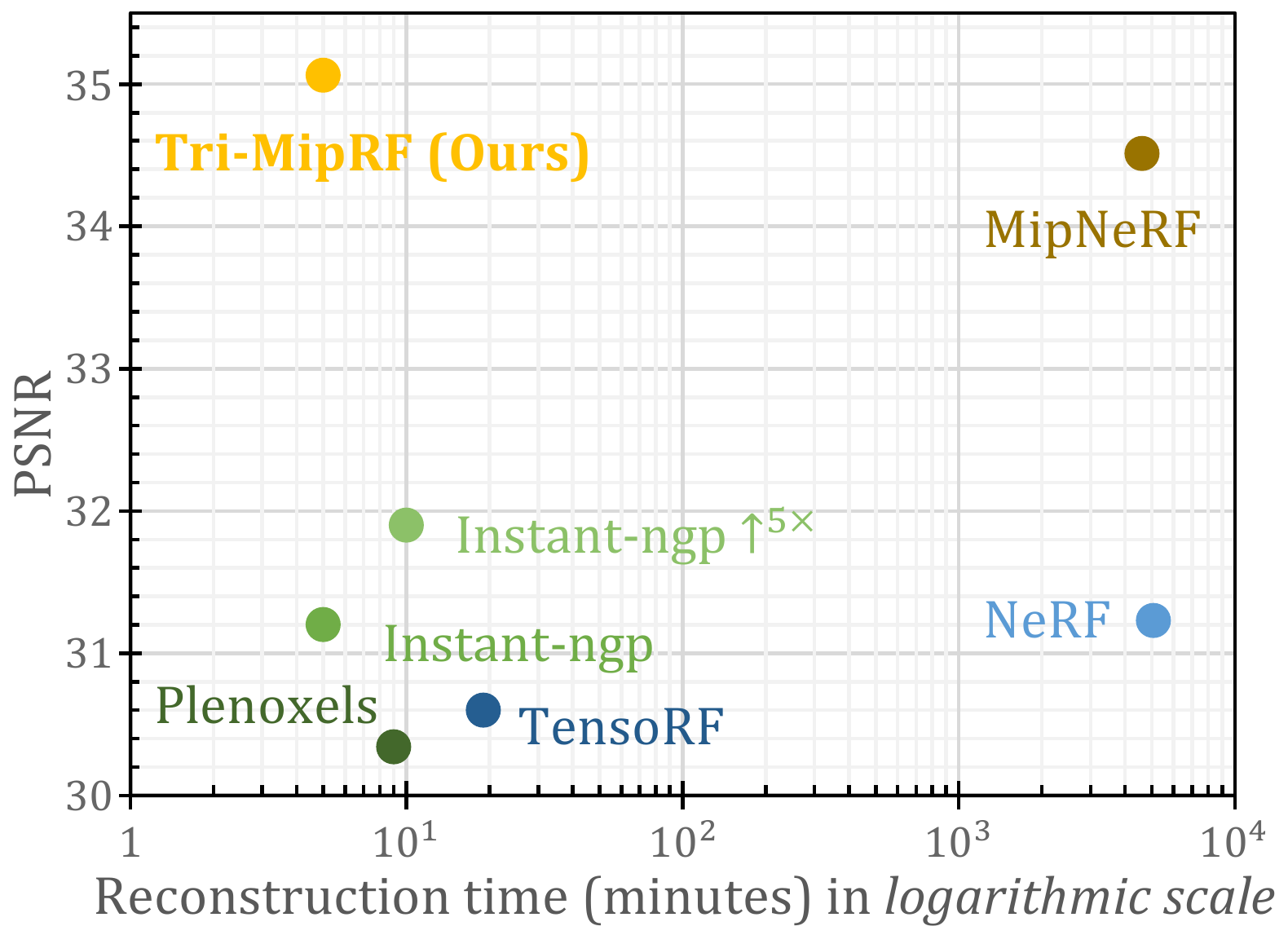}
	\caption{
        Rendering quality \vs reconstruction time on the multi-scale Blender dataset~\cite{barron2021mip}. Our Tri-MipRF achieves state-of-the-art rendering quality while can be reconstructed efficiently, compared with cutting-edge radiance fields methods, \eg, NeRF~\cite{mildenhall2020nerf}, MipNeRF~\cite{barron2021mip}, Plenoxels~\cite{fridovich2022plenoxels}, TensoRF~\cite{chen2022tensorf}, and Instant-ngp~\cite{muller2022instant}. 
        Equipping Instant-ngp with super-sampling (named Instant-ngp $\uparrow^{5\times}$) improves the rendering quality to a certain extent but significantly slows down the reconstruction.
	}
	\vspace{-1.5em}
	\label{fig:teaser}
\end{figure} 

\vspace{-1em}
\section{Introduction}
\label{sec:introduction}

%
Neural radiance field (NeRF)~\cite{mildenhall2020nerf}, emerged as a groundbreaking implicit 3D representation, models the geometry and view-dependent appearance by a multi-layer perceptron (MLP) for rendering photo-realistic novel views.
MipNeRF~\cite{barron2021mip} further pushes the boundaries of rendering quality by integrated position encoding to model the pre-filtered radiance fields.
Such impressive visual quality, however, requires expensive computation in both reconstruction and rendering stages, \eg, MipNeRF~\cite{barron2021mip} takes more than three days for the reconstruction and minutes for rendering a frame.
On the other hand, recent works proposed explicit or hybrid representation for efficient rendering~\cite{reiser2021kilonerf,garbin2021fastnerf,yu2021plenoctrees,hedman2021baking,chen2022mobilenerf,bovzivc2022neural}, or reconstruction~\cite{fridovich2022plenoxels,sun2022direct,chen2022tensorf,muller2022instant}, \eg, the hash encoding~\cite{muller2022instant} greatly reduces the reconstruction time from days to minutes and achieves real-time rendering.
But all their rendering model is flawed in point-based sampling,
which would cause the renderings excessively blurred in close-up views and aliased in distant views.
We face a dilemma of the trade-off between quality and efficiency due to the lacking of a representation to support efficient area sampling.

In this paper, we aim to design a radiance field representation that supports both high-fidelity anti-aliased renderings and efficient reconstruction. 
%
%
To address the aliasing and blurring issue, super-sampling and pre-filtering (\aka area-sampling) are two popular streams of strategies in the offline and real-time rendering literature, respectively.
But super-sampling each pixel by casting multiple rays through its footprint significantly increases the computation cost, and directly pre-filtering the 3D volume is also memory- and computation-intensive, which conflicts with the goal of efficiency.
Also, it is not trivial to pre-filter the radiance field represented with hash encoding, due to the hash collisions.
We achieve this challenging goal with our novel Tri-Mip radiance fields (Tri-MipRF).
As shown in Fig.~\ref{fig:teaser}, our Tri-MipRF achieves state-of-the-art rendering quality that presents high-fidelity details in close-up views and is free of aliasing in distant views.
Meanwhile, it can be reconstructed super-fast, \ie, within five minutes on a single GPU, while the super-sampling variant of hash encoding, Instant-ngp $\uparrow^{5\times}$, takes around ten minutes for the reconstruction and has much lower rendering quality.



The key to achieving our goal is the proposed $\trimip$ encoding, \ie, featurizing the 3D space by three 2D mip (\emph{multum in parvoto}) maps.
The $\trimip$ encoding first decomposes the 3D space into three planes ($\planeXY$, $\planeXZ$, and $\planeYZ$) inspired by the factorization for 3D content generation in~\cite{chan2022efficient}, and then represent each plane by a mipmap.
%
It ingeniously models the pre-filtered 3D feature space by taking advantage of different levels of the 2D mipmaps. 
%
Our Tri-MipRF belongs to the hybrid representation since it models the radiance fields by $\trimip$ encoding and a tiny MLP, which makes it converge fast during the reconstruction.
And the model size of our method is relatively compact since the MLP is very shallow and the $\trimip$ encoding only requires three 2D maps to store the base levels of the mipmaps.
To cope with the $\trimip$ encoding, we propose an efficient cone-casting rendering technique that formulates the pixel as a disc and emits a cone for each pixel.
%
%
%
%
%
%
Different from MipNeRF~\cite{barron2021mip} that samples the cone with multivariate Gaussian, we adopt spheres that are inscribed with the cone.
The spheres are further featurized by the $\trimip$ encoding according to their occupied area.
The reason for doing so is that the features in mipmaps are pre-filtered isotropically.
%
%
%
%
The $\trimip$ encoding models the pre-filtered 3D feature space while the cone-casting is adaptive to the rendering distance and resolution, and they are effectively connected by the occupied area of the sampling sphere, which makes the renderings of our Tri-MipRF free of blurring in close-up views and aliasing in distant views.
%
%
%
%
Besides, we also develop a hybrid volume-surface rendering strategy to 
%
enable
real-time rendering on consumer-level GPUs, \eg, 60 FPS on an Nvidia RTX 3060 graphics card.

%
%


We extensively evaluated our Tri-MipRF on both public benchmarks and images captured in the wild. 
%
%
Both quantitative and qualitative results demonstrate the effectiveness of our method for high-fidelity rendering and fast reconstruction.
%
%
Our contributions are summarized below.
	\vspace{-2mm}
\begin{itemize}
	\item 
        We propose a novel $\trimip$ encoding to model the pre-filtered 3D feature space by leveraging multi-level 2D mipmaps, which enables anti-aliased volume rendering with efficient area sampling.
	\vspace{-2mm}
	\item 
    We propose a new cone-casting rendering technique that efficiently emits a cone for each pixel while gracefully sampling the cone with spheres on the $\trimip$ encoded 3D space.
    
	\vspace{-2mm}
	\item 
  Our method achieves both state-of-the-art rendering quality and reconstruction speed (within five minutes on a single GPU), while still maintaining a compact representation (with a $25\%$ smaller model size than Instant-ngp). 
 Thanks to the hybrid volume-surface rendering strategy, our method also achieves real-time rendering when deploying on consumer-level devices.
\end{itemize}

\begin{figure*}[!t] 
	\centering
	\includegraphics[width=1.0\linewidth]{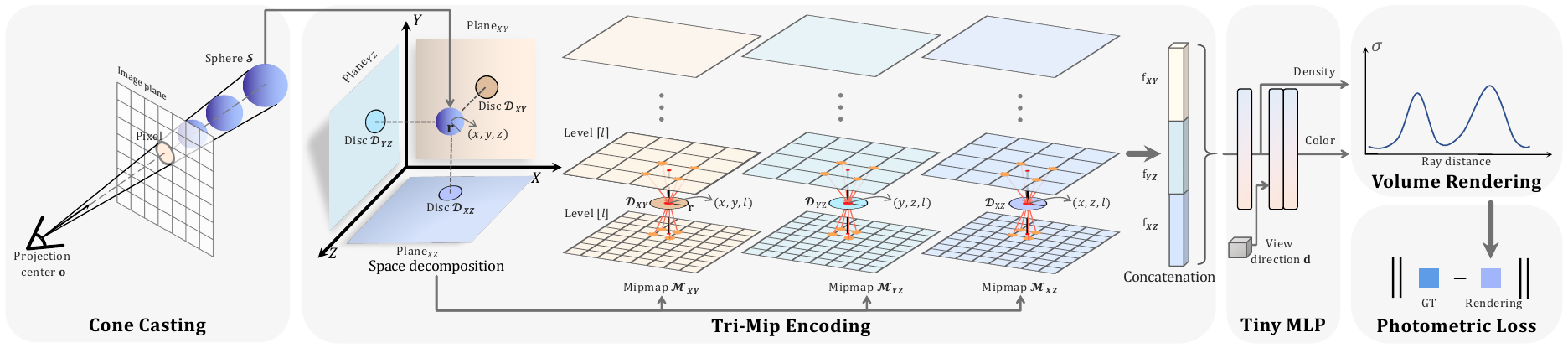}
	\caption{
		Overview of our Tri-MipRF. To render a pixel, we emit a cone from the camera's projection center to the pixel on the image plane, and then we cast a set of spheres inside the cone, next, the spheres are orthogonally projected on the three planes and featurized by our $\trimip$ encoding, after that the feature vector is fed into the tiny MLP to non-linearly map to density and color, finally, the density and color of the spheres are integrated using volume rendering to produce final color for the pixel.
  }
		 
%

	\vspace{-4mm}
	\label{fig:method}
\end{figure*} 

\section{Related Work}
\label{sec:relatedWork}

\myparagraph{Anti-aliasing in rendering.}
Anti-aliasing is a fundamental problem in computer graphics and image processing and has been extensively explored in the rendering community.
Mathematically, aliasing is the effect of overlapping frequency components resulting from an insufficient sampling rate.
Super-sampling and pre-filtering (area-sampling) are two typical streams of approaches to reduce the aliasing artifacts in offline and real-time rendering algorithms, respectively.
Super-sampling anti-aliasing (SSAA) methods~\cite{fuchs1985fast,deering1988triangle,mammen1989transparency,haeberli1990accumulation,whitted2005improved} directly increases the sampling rate to approach the Nyquist frequency, and multi-sampling anti-aliasing (MSAA)~\cite{akeley1993reality} is the de facto method supported by modern graphics processors and APIs.
Pre-filtering-based methods~\cite{10.1145/965105.807509,olano2010lean,kaplanyan2016filtering,akenine2019real,wu2019accurate,kuznetsov2021neumip} relieve this burden by pre-compute the filtered version of content ahead of rendering, thus, this streams of methods are more suitable for real-time rendering.

In the context of NeRF, super-sampling can be achieved by casting multiple rays per pixel and aggregating rendered results to produce the final color.
This straightforward strategy is useful but expensive since the computation cost grows significantly with the sampling rate increasing. 
%
%
On the other hand, recent works~\cite{barron2021mip,barron2022mip,lindell2022bacon} introduce the pre-filtering idea into neural radiance fields by the proposed integrated position encoding or band-limited coordinate networks to learn a pre-filtered representation of the scene, such that the 
renderings of them
are free of blurring in close-up views and aliasing in distant views. 
However, the rendering and reconstruction of them are extremely slow, \eg, MipNeRF~\cite{barron2021mip} takes around three days to reconstruct a scene and minutes to render a frame, which hinders the applicability.
In contrast, our Tri-MipRF can be reconstructed within \emph{five minutes} and achieves real-time rendering on the same hardware, meanwhile, our method even has better-rendering quality in both close-up and distant views compared with MipNeRF.

\myparagraph{Accelerating NeRF.}
NeRF~\cite{mildenhall2020nerf} implicitly represents the scene in the MLP, which leads to a very compact storage, but the reconstruction and rendering of it are extremely slow.
A thread of works is devoted to speeding up the rendering, 
by splitting a scene into many cells~\cite{rebain2021derf,reiser2021kilonerf} to reduce the inference complexity, 
learning to reduce samples per ray~\cite{lindell2021autoint,neff2021donerf}, 
or caching trained fields values~\cite{hedman2021baking,garbin2021fastnerf,yu2021plenoctrees,bovzivc2022neural} to reduce the computation in rendering.
Another line of works focuses on reducing the reconstruction time by directly optimizing the explicit representation~\cite{fridovich2022plenoxels,sun2022direct}, or utilizing hybrid representations, \eg, low-rank tensor~\cite{chen2022tensorf} and hash table~\cite{muller2022instant}, to speed up the converging.
Especially, hash encoding achieves instant reconstruction in around five minutes and rendering in real-time.

However, the rendering model of all the above methods is flawed in formulating the pixel as a single point and sampling with ignorance of the corresponding area, 
which would cause the renderings excessively blurred in close-up views and aliased in distant views.
The super-sampling technique mentioned above can relieve this issue but requires casting multiple rays per pixel, which significantly increases the reconstruction and rendering cost.
And incorporating pre-filtering	with the hash encoding~\cite{muller2022instant} is non-trivial due to the hash collisions.
Our method addresses this issue by the proposed $\trimip$ encoding to effectively model the pre-filtered 3D feature space, which is as efficient as the hash encoding but able to produce anti-aliased high-fidelity renderings.


\myparagraph{Compact 2D representation for 3D content.}
Directly representing 3D contents in volumes is memory- and computation-intensive, as well as redundant since 3D contents are always sparse.
Peng \etal.~\cite{peng2020convolutional} propose to project features of point cloud to multiple planes for 3D geometry reconstruction.
And recent works~\cite{hu-2020-mononizing,wu2021embedding,xia2022lf2mv} have demonstrated that 3D content can be compactly represented in 2D images with faithful restoration.
In the context of generative models, EG3D~\cite{chan2022efficient} proposed a tri-plane representation to decompose 3D volume into three 2D planes for 3D content generation, and this representation is adopted in many follow-up generative methods~\cite{gao2022get3d,shue20223d,singer2023text4d,sun2022ide,bautista2022gaudi,wu2022learning,kangle2023pix2pix3d}.
Besides, this representation is further generalized into 4D space to model dynamic scenes~\cite{cao2023hexplane,fridovich2023k}.
Our $\trimip$ encoding is inspired by this line of works, but none of the above representations can realize our goal, \ie, modeling the pre-filtered 3D feature space for efficient area sampling.

\section{Method}
\label{sec:method}

\subsection{Overview}
\label{subsec:overview}

Given a set of calibrated multi-view images of static scenes, our goal is to efficiently reconstruct the radiance fields that can be further rendered into anti-aliased high-fidelity images.
%
The rendering of radiance fields is performed one pixel at a time, so we describe the rendering procedure of a pixel of interest here, as shown in Fig.~\ref{fig:method}. 
%
We formulate the pixel as a disc on the image plane and perform cone casting for each pixel, rather than ray casting that ignores the area of a pixel.
The cone casting emits a cone $\cone$ from the projection center of the camera to the pixel disc on the image plane, and samples the cone with a set of spheres $\sphere$ that are inscribed with the cone. 
Further, we featurize the spheres to feature vectors $\feat$ by our proposed $\trimip$ encoding that is parameterized by three mipmaps $\mipmap$.
This is the key to making our renderings contain fine-grained details in close-up views and free of aliasing in distant views, since the $\trimip$ encoding effectively models the pre-filtered 3D feature space by taking advantage of different levels in the mipmap.
Then, we employ a tiny MLP parameterized by weights $\modelweights$ to non-linearly map the feature vector $\feat$ of spheres $\sphere$ and view direction $\viewdir$ to density $\density$ and color $\col$ of the spheres,
\begin{gather}
\label{eq:mlp}
	[\density,\col] = \mlp(\feat, \viewdir; \; \modelweights).
\end{gather}
Finally, the estimated densities and colors of spheres inside a cone are used to approximate the volume rendering integral by numerical quadrature as in~\cite{max1995optical} to render the final color of the pixel corresponding to the cone:
\begin{gather}
\Col(\zvec,\viewdir, \modelweights, \mipmap) = \sum_{i} T_i \left(1-\exp(-\density_i (t_{i+1} - t_i)) \right) \col_i,  \nonumber
\\
\textrm{with}\quad T_i = \exp \bigg(\!-\! \sum_{k < i}  \density_{k} \left(t_{k+1} - t_{k} \right)\!\bigg),
\end{gather}
where $\zvec$ is the distance between the sampled spheres and the projection center of the camera. 
During training, the photometric loss will be computed between the rendered colors and captured colors to back-propagate gradients to the weights $\modelweights$ of MLP and parameters $\mipmap$ of $\trimip$ encoding to jointly optimize them.

In the following sections, we will present the cone casting, $\trimip$ encoding, as well as the hybrid volume-surface rendering in detail, while omitting the procedures of tiny MLP and volume rendering as they are similar to the original NeRF~\cite{mildenhall2020nerf}. \supp{Please refer to the supplemental material for more details.}

\subsection{Cone Casting}
\label{subsec:cone_casting}

\begin{figure}[!t] 
	\centering
	\includegraphics[width=1.0\linewidth]{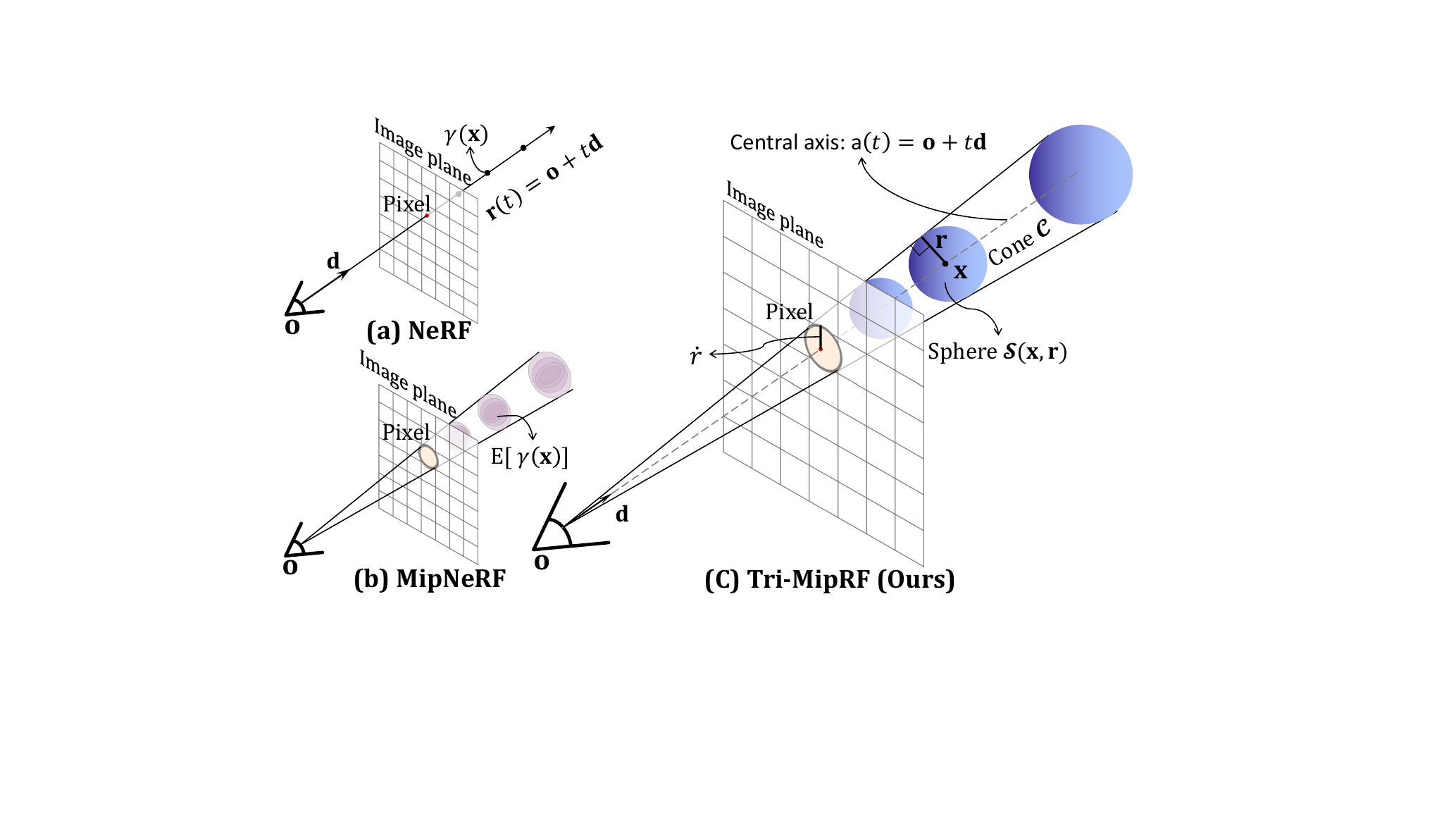}
	\caption{
		NeRF~\cite{mildenhall2020nerf} renders a pixel by ray-casting and the sampling points $\position$ on the ray are featurized by position encoding $\posenc$. MipNeRF~\cite{barron2021mip} emits a cone for each pixel and featurize the sampling multivariate Gaussian by integrated position encoding $\operatorname{E}[\posenc]$. Our Tri-MipRF renders a pixel by cone casting and the cone is sampled by a set of spheres that are inscribed with the cone.
	}
		\vspace{-4mm}
	\label{fig:cone}
\end{figure}

NeRF renders a pixel by emitting a ray $\ray(t) = \rayorigin + t \raydir$, and sampling points $\position$ along the ray, \aka ray casting, as shown in Fig.~\ref{fig:cone} (a).
And the points $\position$ are further featurized by position encoding (PE) $\gamma$ to produce the feature vectors for the points $\posenc$.
This formulation models the pixel as a single point while ignoring the area of the pixel, which is quite different from the real-world imaging sensors.
Most NeRF works~\cite{sun2022direct,yu2021plenoctrees,fridovich2022plenoxels,chen2022tensorf,chen2022mobilenerf}, including instant-ngp~\cite{muller2022instant}, followed this formulation.
It can approximate the real-world case when the captured/rendered views are at a roughly constant distance but will lead to obvious artifacts when viewing at very different distances, \eg, blurring in close-up views and aliasing in distant views since the sampling is distance-agnostic.
To this end, MipNeRF emits a cone for each pixel and samples the cone by the multivariate Gaussian, which is further featurized by integrated position encoding (IPE). 
The IPE is derived by the integral $\operatorname{E}[\posenc]$ over the PE of the points within the Gaussian, as shown in Fig.~\ref{fig:cone} (b).
%
This strategy, however, is not trivial to be extended to explicit or hybrid representations for efficient reconstruction and rendering, \eg, hash encoding~\cite{muller2022instant}, 
since IPE is the integral of coordinate-based positional encoding, which is not compatible with explicit or hybrid volumetric feature encoding.

In contrast, our efficient cone-casting strategy can efficiently work with our $\trimip$ encoding for area sampling during the volume rendering. 
%
As shown in Fig.~\ref{fig:cone} (c), we formulate the pixel as a disc on the image plane rather than a single point that ignores the area of a pixel.
The radius of the disc can be calculated by $\baseradius = \sqrt{\nicefrac{\Delta x \cdot \Delta y}{\pi}}$, where $\Delta x$ and $\Delta y$ are the width and height of the pixel in world coordinates that can be derived from the calibrated camera parameters.
For each pixel, we emit a cone $\cone$ from the camera's projection center $\rayorigin$ along the direction $\raydir = \pixcenter - \rayorigin$, where $\pixcenter$ is the pixel's center.
The apex of the cone is located at the optical center of the camera and the intersection between the cone and the image plane is the disc corresponding to the pixel.
We can derive the central axis of the cone as $\mathbf{a}(t) = \rayorigin + t\raydir$.
%
%
%
To sample the cone, we cannot follow MipNeRF~\cite{barron2021mip} to use the multivariate Gaussian, since the multivariate Gaussian is anisotropic but the pre-filtering in our $\trimip$ encoding is isotropic. 
Thus, we sample the cone with a set of spheres $\sphere(\position, \; \radius)$ parameterized by their centers $\position$ and radiuses $\radius$.
%
The centers $\position$ are located at the central axis of the cone and the radiuses $\radius$ are set to make the spheres inscribed with the cone, which can be written as:
%
\begin{equation}
\begin{aligned}
\label{eq:sphere}
	\position &= \rayorigin + t \raydir \,, 
	\\
	\radius &= \frac{
		\norm{\position-\rayorigin}_2 \cdot f\baseradius
		}
		{
		\norm{\raydir}_2 \cdot \sqrt{
			\bigg(
			\sqrt{\norm{\raydir}_2^2 - f^2} - \baseradius
			\bigg)^2 + f^2
			},
		}
\end{aligned}
\end{equation}
where $f$ is the focal length.
Based on Eq.~\ref{eq:sphere}, the sampling spheres $\sphere(\position, \; \radius)$ can be determined by a sorted distance vector $t_i \in \zvec$, since the center location $\position_i$ and radius $\radius_i$ of a sphere $\sphere_i$ is the function of the distance $t_i$. 
We uniformly sample $t_i \in \zvec$ between the camera's predefined near $t_n$ and far $t_f$ planes or the two intersections between the central axis of the cone and the axis-aligned bounding-box (AABB) of the interested 3D space.
To further speed up the cone casting by utilizing the sparsity of the 3D space, we employ a binary occupancy grid that coarsely marks empty \vs. non-empty space similar to~\cite{muller2022instant,li2022nerfacc}, such that we can cheaply skip samples in the empty area and concentrate samples near surfaces to avoid wasted computation.

\subsection{Tri-Mip Encoding}
\label{subsec:trimip}

To realize our goal, \ie rendering fine-grained details in close-up views and avoiding aliasing in distant views while maintaining the reconstruction and rendering efficiency, we should constructively featurize the sampled spheres $\sphere(\position, \; \radius)$ according to their occupied area, which shares similar motivation of area-sampling (\aka pre-filtering) in computer graphics.
Hash encoding proposed in instant-ngp~\cite{muller2022instant} can efficiently featurize the sampled \emph{points} by looking up the hash table and trilinear interpolation, however, it cannot be easily extended to featurize the spheres $\sphere(\position, \; \radius)$.
%
One plausible workaround is to
incorporate the super-sampling strategy with hash encoding to approximate the featurization of spheres.
%
However, super-sampling
significantly increases the computation cost, which
unexpectedly sacrifices the ability of
efficient reconstruction and rendering. 

%

To this end, we propose a novel $\trimip$ encoding parameterized by three trainable mipmaps $\mipmap$ to featurize the sampling spheres $\sphere(\position, \; \radius)$:
\begin{gather}
\begin{aligned}
\label{eq:trimip}
	\feat &= \trimip(\position, \radius; \; \mipmap),
	\\
	\mipmap &= \{\mipmap_{XY}, \mipmap_{XZ},\mipmap_{YZ}\}.
\end{aligned}
\end{gather}
As shown in Fig.~\ref{fig:method}, the $\trimip$ encoding decomposes the 3D space into three planes ($\planeXY$, $\planeXZ$, and $\planeYZ$) using orthographic projection, and then represent each plane by a mipmap ($\mipmap_{XY}$, $\mipmap_{XZ}$, and $\mipmap_{YZ}$) to model the pre-filtered feature space.
For each mipmap, the base level $\mipmap^{L_0}$ is a feature map with the shape of $H \times W \times C$, where $H, W, C$ are the height, width, and number of channels, respectively.
The base level $\mipmap^{L_0}$ is randomly initialized and can be trained during the reconstruction, and other levels ($\mipmap^{L_i}, i={1,2,...,N}$) are derived from the previous level $\mipmap^{L_{i-1}}$ by downscaling $2\times$ along the height and width.
This pre-filtering strategy maintains consistency among the levels of mipmap, which is the key to making the reconstructed objects coherent at different distances.

To query the feature vectors $\feat$ corresponding to the spheres $\sphere(\position, \; \radius)$, we first orthogonally project $\sphere$ on the three planes to obtain three discs $\disc = \{\disc_{XY}, \disc_{XZ}, \disc_{YZ}\}$, as shown in Fig.~\ref{fig:method}.
For each disc, we query a feature vector from the corresponding mipmap.
Take disc $\disc_{XY}$ as an example, we query its feature $\feat_{XY}$ from the mipmap $\mipmap_{XY}$.
%
%
Based on the property of orthogonal projection, the disc $\disc_{XY}$ shares the same radius $\radius$ as the sampled sphere, and the 2D coordinate of the $\disc_{XY}$'s center $\position_{\disc_{XY}}$ is the partial coordinate $(x,y)$ of the sampled sphere's center $\position(x,y,z)$.
%
%
For the disc $\disc_{XY}$'s query level $l$ of the mipmap $\mipmap_{XY}$, we assign it to:
 \begin{equation}
 	\begin{aligned}
 	\label{eq:level}
 		l &= log_2\bigg(\frac{\radius}{\featradius}\bigg), 
 		\\
 		\featradius &= \sqrt{\frac{(\aabb_{max}-\aabb_{min})_X \cdot (\aabb_{max}-\aabb_{min})_Y}{HW\cdot\pi}},
 	\end{aligned}
 \end{equation}
where $\featradius$ is the radius of the feature elements in the base level of the mipmap $\mipmap^{L_0}$, $\aabb_{max}$ and $\aabb_{min}$ are the maximum and minimum corners of the Axis Aligned Bounding Box (AABB) of the interested 3D space, respectively.
%
The motivation of Eq.~\ref{eq:level} is to match the sphere's radius $\radius$ with the feature elements' radius in a certain level of the mipmap $\mipmap_{XY}^l$.
After obtaining the query coordinate $(x,y,l)$, we can get the feature vector $\feat_{XY}$ from the mipmap $\mipmap_{XY}$ by the trilinear interpolation.
As shown in Fig.~\ref{fig:method}, we first find the two nearest levels of the mipmap $\mipmap_{XY}^{\lfloor l \rfloor}$ and $\mipmap_{XY}^{\lceil l \rceil}$; and then we project the center coordinate $(x,y)$ of the disc $\disc_{XY}$ to the two levels of the mipmap (shown as red dots); next, we find four neighbors of them (shown as orange dots), respectively; finally, we interpolate the eight neighbors based on their distance to the center of the disc $\disc_{XY}$ to produce the feature vector $\feat_{XY}$.
%
The trilinear interpolation increases the effective precision of both levels and spatial resolutions, also, it yields a continuous encoding space that is beneficial for efficient training.
Similarly, we can get the feature vectors $\feat_{XZ}$ and $\feat_{YZ}$ for the disc $\disc_{XZ}$ and $\disc_{YZ}$, respectively.
The final queried feature vector $\feat$ for the sampled sphere $\sphere$ is a concatenation of the three discs' feature vectors $\{\feat_{XY}, \feat_{XZ}, \feat_{YZ}\}$.

Our $\trimip$ encoding effectively featurize the 3D space in a pre-filtered way, such that we can perform area-sampling for the volume rendering to produce high-quality renderings that are free of aliasing.
And the feature query process is also efficient, \ie, querying mipmap has been highly optimized in modern GPUs, which promotes fast reconstruction.
Besides, the storage of our $\trimip$ encoding is 
three 2D feature maps, \ie the base levels of the three mipmaps $\mipmap^{l_0}$ as other levels are derived by the base level by downscaling, which 
makes our model compact enough for easy distribution.
%
%
Note that, $\trimip$ encoding also promotes the training converges faster than implicitly representing the scene in MLP, \eg, our method only takes $25\thousand$ iterations 
to converge while MipNeRF~\cite{barron2021mip} requires $1\million$ iterations, since features in the mipmap $\mipmap$ can be optimized directly rather than mapped from the IPE by optimizable weights of MLP. 

\begin{figure}[!t] 
	\centering
	\includegraphics[width=1.0\linewidth]{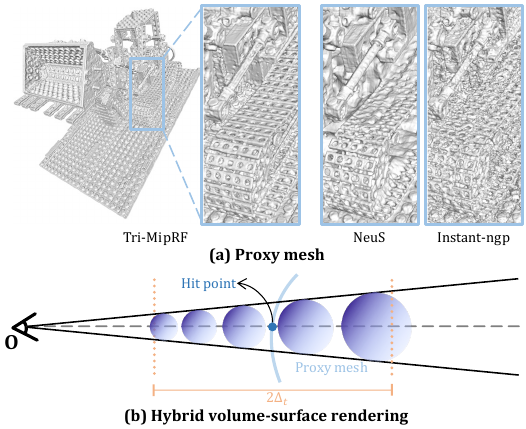}
	\caption{
	Visual comparison of the proxy mesh produced by our Tri-MipRF, Instant-ngp~\cite{muller2022instant}, and NeuS~\cite{wang2021neus} (a); and our proposed hybrid volume-surface real-time rendering strategy (b).
	}
		\vspace{-4mm}
	\label{fig:rendering}
\end{figure} 

\begin{table*}[]
    \renewcommand{\tabcolsep}{1pt}
    \centering
    \resizebox{\linewidth}{!}{
    \begin{tabular}{@{}l@{\,\,}|l@{\,}|ccccc|ccccc|ccccc}
    &   & \multicolumn{5}{c|}{PSNR $\uparrow$} & \multicolumn{5}{c|}{SSIM $\uparrow$} & \multicolumn{5}{c}{LPIPS $\downarrow$}  \\
    & Train. $\downarrow$  & Full Res. & $\nicefrac{1}{2}$ Res. & $\nicefrac{1}{4}$ Res. & $\nicefrac{1}{8}$ Res. & Avg. & Full Res. & $\nicefrac{1}{2}$ Res. & $\nicefrac{1}{4}$ Res. & $\nicefrac{1}{8}$ Res. & Avg. & Full Res. & $\nicefrac{1}{2}$ Res. & $\nicefrac{1}{4}$ Res. & $\nicefrac{1}{8}$ Res & Avg.  \\ \hline
NeRF w/o $\mathcal{L}_\text{area}$ & 3 days  &                    31.20 &                    30.65 &                    26.25 &                     22.53 &  27.66 &                   0.950 &                    0.956 &                    0.930 &                    0.871 &  0.927 &                  0.055 &                    0.034 &                   0.043 &                    0.075 &                    0.052
\\
NeRF~\cite{mildenhall2020nerf}  & 3 days &                    29.90 &                    32.13 &                    33.40 &                     29.47 &  31.23 &                   0.938 &                    0.959 &                    0.973 &                    0.962 &  0.958 &                  0.074 &                    0.040 &                   \cellcolor{yellow} 0.024 &                    0.039 &                  0.044 
\\
MipNeRF~\cite{barron2021mip}   & 3 days & \cellcolor{orange} 32.63 &     \cellcolor{orange}34.34 & \cellcolor{orange}35.47 & \cellcolor{orange}35.60 & \cellcolor{orange}34.51 & \cellcolor{orange}0.958 & \cellcolor{orange}0.970 & \cellcolor{orange}0.979 & \cellcolor{orange}0.983 & \cellcolor{orange}0.973& \cellcolor{orange}0.047 & \cellcolor{orange}0.026 & \cellcolor{red}0.017 & \cellcolor{orange}0.012 & \cellcolor{orange}0.026 
\\
\hline
Plenoxels~\cite{fridovich2022plenoxels}  & 9 min  &  31.60 & 32.85 & 30.26 & 26.63 & 30.34 & \cellcolor{yellow} 0.956 &  \cellcolor{yellow}0.967 & 0.961 & 0.936 & 0.955 &  \cellcolor{yellow}0.052 &  \cellcolor{yellow} 0.032 &  0.045 & 0.077 & 0.051
\\
TensoRF~\cite{chen2022tensorf}  & 19 min  &    \cellcolor{yellow}32.11 &                    \cellcolor{yellow}33.03 & 30.45 &  26.80 &                    30.60 & \cellcolor{yellow} 0.956 &  0.966 &  0.962 &                    0.939 &                    0.956 & 0.056 & 0.038 &0.047 & 0.076 & 0.054 
\\
Instant-ngp~\cite{muller2022instant}   &\cellcolor{orange}5 min  &     30.00 & 32.15 &                    33.31 &                    29.35 & 31.20 &   0.939 &                    0.961 &  0.974 &  0.963 &0.959 &                    0.079 & 0.043 &  0.026 & 0.040 &0.047  
\\
Instant-ngp $\uparrow^{5\times}$    &10 min&     30.96 & 32.87 &                    33.10 &                    \cellcolor{yellow}30.82 & \cellcolor{yellow}31.94 &   0.945 &                    0.965 & 0.973 & \cellcolor{yellow}    0.970 & \cellcolor{yellow}    0.963 &                    0.070 & 0.038 &  0.025 & \cellcolor{yellow}0.029 &\cellcolor{yellow}0.041  
\\
\hline
Tri-MipRF w/o $\mipmap$       &\cellcolor{red}4.5 min    &                                        30.25 &                    32.52 &                    \cellcolor{yellow}33.73 &                    29.44 &    31.48 &                0.938 &                    0.961 &                    \cellcolor{yellow}0.975 &                    0.964 &                    0.959 &                    0.081 &                    0.045 &                    0.026 & 					0.039 &		 
0.048	
\\
Tri-MipRF (Ours)        &\cellcolor{orange}5 min    &                    \cellcolor{red}33.32 &                    \cellcolor{red}35.02 &                    \cellcolor{red}35.78 &                    \cellcolor{red}36.13 &                    \cellcolor{red}35.06 &                    \cellcolor{red}0.961 &                    \cellcolor{red}0.974 &                    \cellcolor{red}0.981 &                    \cellcolor{red}0.986 &                    \cellcolor{red}0.976 &                    \cellcolor{red}0.043 &                    \cellcolor{red}0.024 &                    \cellcolor{red}0.017 & 					\cellcolor{red}0.011 &					  \cellcolor{red}0.024
    \end{tabular}
    }
    \vspace{0.1em}
    \caption{
    Quantitive comparison of our Tri-MipRF against several cutting-edge methods and their variants on the multi-scale Blender dataset.  
    }
    \vspace{-1em}
    \label{tab:avg_multiblender_results}
\end{table*}

\subsection{Hybrid Volume-Surface Rendering}
\label{subsec:rendering}
Though our method can efficiently reconstruct the radiance fields, directly rendering it on consumer-level GPUs, \eg, an Nvidia RTX 3060 graphics card, only achieves around 30 FPS.
This is because the volume rendering inherently samples multiple spheres inside the cone for each pixel, though we can skip some samples by the binary occupancy grid.
%
Observing the real-time surface rendering benefited from the efficient rasterization of the polygon mesh, we develop a hybrid volume-surface rendering strategy to further boost the rendering speed.
Besides the reconstructed radiance field, our hybrid volume-surface rendering strategy requires a proxy mesh to efficiently determine a rough distance from the camera's optical center to the object.
Fortunately, we can obtain the proxy mesh by marching cubes~\cite{lorensen1987marching} on the reconstructed density field followed by mesh decimation.
The proxy mesh produced by our Tri-MipRF presents high-fidelity quality even in complicated structure details, as shown in the left-hand side of Fig.~\ref{fig:rendering} (a), while the results produced by Instant-ngp~\cite{muller2022instant} and NeuS~\cite{wang2021neus} are shown in the right-hand side as references.

Once the proxy mesh is available,
we first efficiently rasterize it to obtain the hit point (shown as a blue dot) on the surface for the central axis of the cone, as shown in Fig.~\ref{fig:rendering} (b), then we uniformly sample spheres within the distance of $\Delta_t$ from the hit point in the central axis of the cone, which yields a $2\Delta_t$ sampling interval.
This hybrid volume-surface rendering strategy significantly reduces the number of samples, thus, enabling real-time rendering (>60 FPS) on consumer-level GPUs.
\supp{
Please refer to the video in the supplemental material for the real-time interactive rendering demo.
}
\section{Experimental Evaluation}
\label{sec:experiment}

\subsection{Implementation}
Our Tri-Mip radiance fields (Tri-MipRF) learns the 3D structure solely from the calibrated multi-view 2D images and is trained with the photometric metric loss between the rendered pixels and the captured ones.
Following MipNeRF~\cite{barron2021mip}, we scale the loss of each pixel by the area of its footprint on the image plane, noted as ``area loss $\mathcal{L}_\text{area}$''.
We implement our Tri-MipR using PyTorch~\cite{paszke2019pytorch} framework with tiny-cuda-nn~\cite{tiny-cuda-nn} extension. 
%
The mipmap query process is well optimized in the rendering community for texture sampling, thus, we employ the nvdiffrast~\cite{Laine2020diffrast} library to implement our $\trimip$ encoding efficiently.
The shape of the base level of the mipmap $\mipmap^{l_0}$ in $\trimip$ encoding is empirically set to $H=512,W=512,C=16$, which is a compact representation for the scene.
We train our Tri-MipRF using the AdamW optimizer~\cite{loshchilov2019decoupled} for $25\thousand$ iterations with the weight decay set to $1\times10^{-5}$ and the learning rate set to $2\times10^{-3}$ and scheduled by MultiStepLR in PyTorch.
Note, the learning rate for the $\trimip$ encoding is further scaled up $10\times$ since the parameter $\mipmap$ of $\trimip$ encoding directly represents the scene while that for tiny MLP keeps unchanged.

\subsection{Evaluation on the Multi-scale Blender Dataset}
\label{subsec:ms-blender}

The Blender dataset presented in the original NeRF~\cite{mildenhall2020nerf} is a synthetic dataset where all training and testing images observe the scene content from a roughly constant distance, which is very different from real-world captures.
MipNeRF~\cite{barron2021mip} presents a multi-scale Blender dataset to better probe the reconstruction accuracy and anti-aliasing on multi-resolution scenes.
It is compiled by downscaling the original dataset with a factor of 2, 4, and 8, and combining them together. 
Due to the nature of projecting geometry, this is almost equivalent with re-rendering the original dataset where the distance to the camera has been increased by scale factors of 2, 4, and 8.

\begin{figure*}[!t] 
	\centering
	\includegraphics[width=0.95\linewidth]{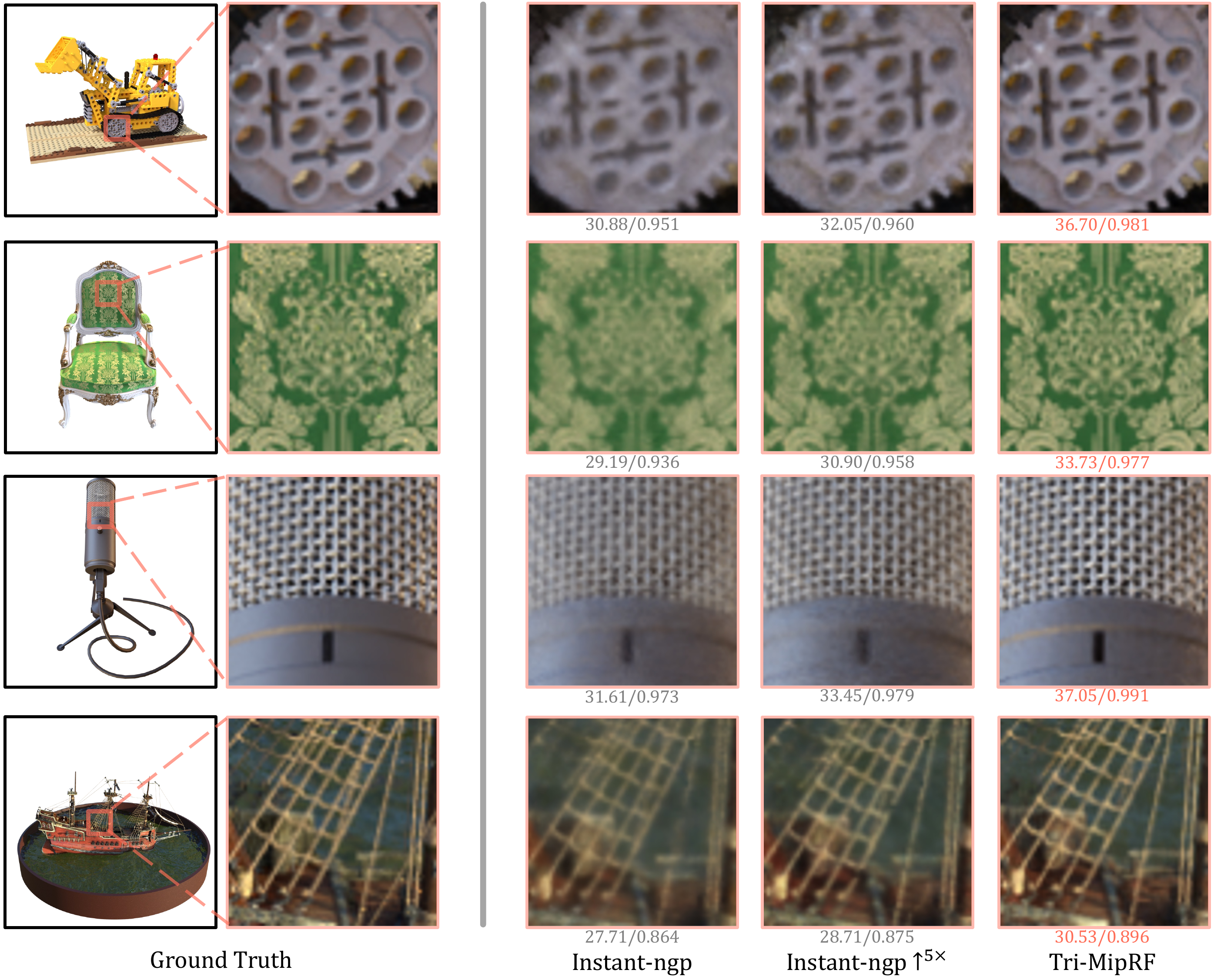}
	\vspace{-0.5em}
	\caption{
		Qualitative comparison of the full-resolution (close-up views) renderings on the multi-scale Blender dataset. PSNR/SSIM values are shown at the bottom of each result.
	}
    \vspace{-1em}
	\label{fig:qualitative_multiscale}
\end{figure*} 

\myparagraph{Quantitative results.}
We compared our Tri-MipRF with several cutting-edge methods, \ie, NeRF~\cite{mildenhall2020nerf}, MipNeRF~\cite{barron2021mip}, Plenoxels~\cite{fridovich2022plenoxels}, TensoRF~\cite{chen2022tensorf}, and Instant-ngp~\cite{muller2022instant}.
Following previous works, we report three metrics: PSNR, SSIM~\cite{wang2004image}, and VGG LPIPS~\cite{zhang2018unreasonable}, as shown in Tab.~\ref{tab:avg_multiblender_results}.  
We also report the rough reconstruction time on the same hardware, \ie a single Nvidia A100 GPU. 
%
%
Except for MipNeRF, other comparison methods are not designed for multi-scale captures or imaging at various distances, thus, we equipped all of them with the  aforementioned area loss $\mathcal{L}_\text{area}$ by default, which yields a better performance as evidenced by comparing the results of ``NeRF w/o $\mathcal{L}_\text{area}$'' and ``NeRF''. 
From Tab.~\ref{tab:avg_multiblender_results}, we can see that MipNeRF presents high-quality renderings, however, the reconstruction of it is extremely slow (up to around three days) which greatly prevents the applicability.
Besides, the reconstruction times of Plenoxels, TensoRF, and Instant-ngp are greatly faster than that of MipNeRF, but the rendering qualities are unsatisfactory no matter in terms of PSNR, SSIM, or LPIPS.
For Instant-ngp, we further design a super-sampling variant of it, Instant-ngp $\uparrow^{5\times}$, which means casting five rays in the quincunx sample pattern for each pixel and aggregating the samples in these rays.
%
%
We can find that super-sampling makes it render higher-quality images, however, super-sampling also significantly increases the reconstruction time from five to ten minutes. 
In contrast, our Tri-MipRF not only produces the highest-quality renderings for all four types of resolutions but also can be reconstructed super-fast, \ie 5 minutes.
To verify the effectiveness of the $\trimip$ encoding, we also evaluate an ablation of our method, Tri-MipRF w/o $\mipmap$, that replaces the three mipmaps with three 2D feature maps with the same shape as the base level of the mipmap.
As shown in Tab.~\ref{tab:avg_multiblender_results}, the Tri-MipRF w/o $\mipmap$ performs comparable with Instant-ngp but significantly worse than our full method, Tri-MipRF, even though it can be reconstructed slightly faster than Tri-MipRF since it gets rid of the mipmap query procedure.
These quantitative comparisons demonstrate the effectiveness of our $\trimip$ encoding and cone casting, such that our Tri-MipRF can effectively model the pre-filtered 3D feature space and efficiently perform area sampling on it for anti-aliased high-fidelity rendering.

\begin{figure}[!t] 
	\centering
	\includegraphics[width=\linewidth]{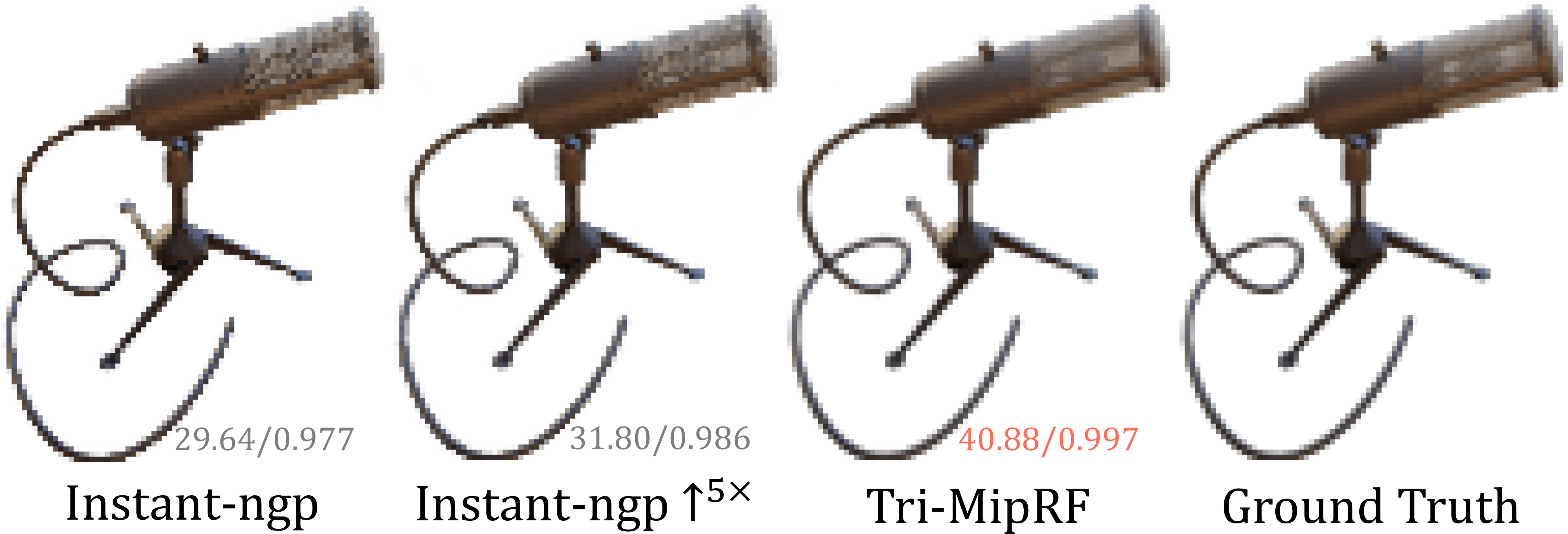}
	\caption{
		Qualitative comparison of the low-resolution renderings (distant view) on the multi-scale Blender dataset. PSNR/SSIM values are shown in the bottom right corners of each result.
	}
		\vspace{-2em}
	\label{fig:qualitative_aliasing}
\end{figure} 

\myparagraph{Qualitative results.}
We further qualitatively compared our Tri-MipRF with the Instant-ngp~\cite{muller2022instant} and its super-sapling variant, Instant-ngp $\uparrow^{5\times}$, since their reconstruction speed is similar to ours, \ie, five minutes for our Tri-MipRF, five and ten minutes for Instant-ngp and Instant-ngp $\uparrow^{5\times}$, respectively.
%
%
In Fig.~\ref{fig:qualitative_multiscale}, we show examples of full-resolution renderings that can be treated as close-up views.
we can see that the results of Instant-ngp suffer from blurriness for structure and texture details, Instant-ngp $\uparrow^{5\times}$ improves the quality but significantly increases the reconstruction time.
In contrast, our method faithfully renders the fine-grained details while keeping the reconstruction super-fast.
On the other hand, we compare the renderings of $\nicefrac{1}{8}$ resolution that can simulate the distant views in Fig.~\ref{fig:qualitative_aliasing}.
We can see renderings of Instant-ngp exhibit severe aliasing and ``jaggies'' artifacts and Instant-ngp $\uparrow^{5\times}$ slightly relieves this issue, while our Tri-MipRF faithfully renders smooth appearance and fine-grained structure details, thanks to the $\trimip$ encoding that efficiently models the pre-filtered 3D feature space.
\supp{We highly recommend readers to watch the supplemental video to better evaluate the anti-aliasing feature.}

\begin{figure*}[!t] 
	\centering
	\includegraphics[width=0.95\linewidth]{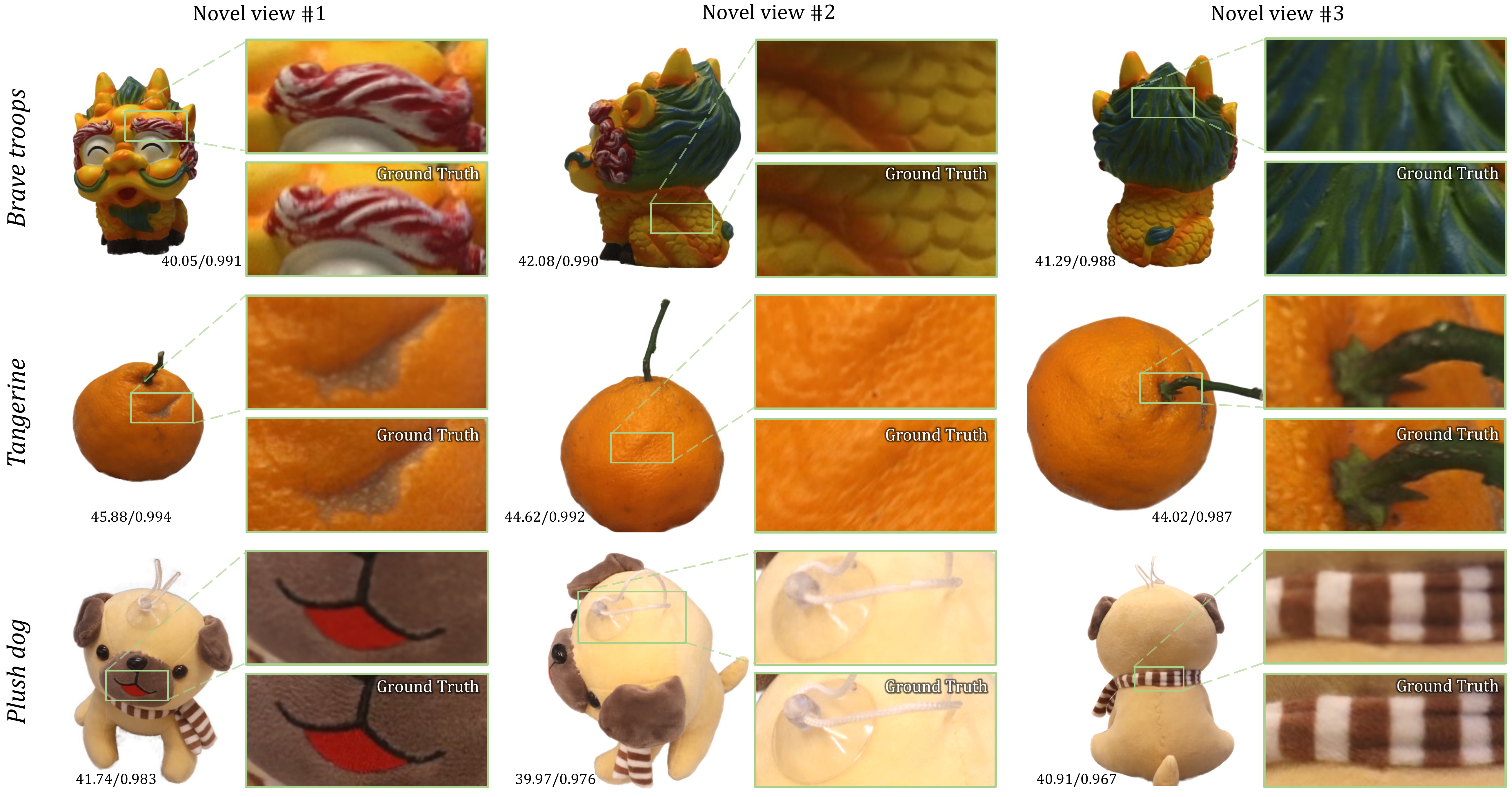}
	\caption{
		Example rendering results of our method from in-the-wild captures. The PSNR/SSIM values are shown below the renderings of our method.
	}
		\vspace{-1.5em}
	\label{fig:in_the_wild}
\end{figure*} 

\subsection{Evaluation on the Single-scale Blender Dataset}
\label{subsec:blender}

The easier single-scale Blender dataset captures images at a roughly constant distance, which is friendly to the point-sampling-based methods, \eg, NeRF~\cite{mildenhall2020nerf}, Plenoxels~\cite{fridovich2022plenoxels}, TensoRF~\cite{chen2022tensorf}, Instant-ngp~\cite{muller2022instant}, and \etc.
We also compared our Tri-MipRF with multiple cutting-edge methods on this dataset, as shown in Tab.~\ref{tab:avg_blender_results}.
We can see that our Tri-MipRF still outperforms all of them no matter in terms of PSNR, SSIM, or LPIPS, in the meanwhile, achieving the fastest reconstruction together with Instant-ngp.
Besides, we also report the model size, \ie the storage consumption, in Tab.~\ref{tab:avg_blender_results}.
We can find that the implicit methods have extremely small model sizes, \eg, the model size of NeRF and MipNeRF is 5.00 MB and 2.50 MB, respectively, but are reconstructed very slowly ($\sim3$ days); the model sizes of explicit methods, \eg, Plenoxels and DVGO, are very large ($>$ 500 MB); and the hybrid methods, \eg, Instant-ngp, TensoRF, and our Tri-MipRF, have relative small model size ($<$ 100 MB) while our Tri-MipRF has the smallest model size (48.2 MB) in the hybrid methods, which reduces $25\%$ storage consumption compared against Instant-ngp.
%
\supp{Please refer to the supplemental materials for more detailed statistics.}

\begin{table}[]
    \renewcommand{\tabcolsep}{6pt}
    \centering
    \resizebox{\linewidth}{!}{
    \begin{tabular}{@{}l|ccc|c@{\,}}
     & \!PSNR $\uparrow$\! & \!SSIM $\uparrow$\! & \!LPIPS $\downarrow$\!   & \# Size  $\downarrow$\!
    \\ 
    \hline
    SRN~\cite{srn}                                                    &                    22.26  &                    0.846  &                    0.170   &  -  \\
LLFF~\cite{mildenhall2019local}                                        &                    24.88  &                    0.911  &                    0.114  &  -  \\
Neural Volumes~\cite{neuralvolumes}                               &                    26.05  &                    0.893  &                    0.160 &  -  \\
Plenoxels~\cite{fridovich2022plenoxels} &                    31.71 &                    0.958 &                    0.049                     & 778 MB \\
NeRF~\cite{mildenhall2020nerf} &                    31.74 &                    0.953 &                    0.050                     & \cellcolor{orange}5.00 MB \\
DVGO~\cite{sun2022direct} & 31.95 & 0.957 & 0.053  & 612 MB \\
MipNeRF~\cite{barron2021mip}      &     33.09 & 0.961 &  \cellcolor{orange}0.043  & \cellcolor{red} 2.50 MB \\
TensoRF~\cite{chen2022tensorf} &                    \cellcolor{yellow}33.14 &                    \cellcolor{red}0.963 &                    0.047                     & 71.8 MB \\
Instant-ngp~\cite{muller2022instant} &                    \cellcolor{orange}33.18 &                    \cellcolor{red}0.963 &                   \cellcolor{yellow}0.045                     & 64.1 MB \\
\hline
Tri-MipRF &                    \cellcolor{red}33.65 &                    \cellcolor{red}0.963 &                    \cellcolor{red}0.042                     & \cellcolor{yellow}48.2 MB 


    \end{tabular}
    }
    \vspace{0.1em}
    \caption{
    Results on the single-scale Blender dataset of our Tri-MipRF and several cutting-edge methods.
    }
    \vspace{-1em}
    \label{tab:avg_blender_results}
\end{table}

\subsection{Applicability on the In-the-wild Captures}
\label{subsec:in-the-wild}

To further demonstrate the applicability of our method, we captured several objects in the wild.
We performed SFM on the sequence to estimate the camera's intrinsic and extrinsic parameters and employed multi-view segmentation methods to separate the object from the background scenes.
Each captures contain $200 \sim 300$ images with the resolution of $1200\times800$, and we uniformly sample $70\%$ of them for the reconstruction and the remains are used for evaluation. 
We show three example results in Fig.~\ref{fig:in_the_wild}, where we can see the rendered novel views faithfully reproduce the detailed structures and appearances, and the PSNR/SSIM values marked below the images also evidence the applicability of our method.
Interestingly, we find our renderings even have ``better'' details than the ground truth in some cases, \eg, the brave troops' eyebrow shown in the blow-up figure of the novel view $\#1$ in the first line of Fig.~\ref{fig:in_the_wild}.  
This is because the ground truth, \ie the captured image by the camera in the wild, may suffer from motion blur artifacts due to the fast movements, while this issue is relieved during the reconstruction by fusing multiple observations.




\section{Conclusion}
\label{sec:conclusion}
In this work, we propose a Tri-Mip radiance fields, Tri-MipRF, to make the renderings contain fine-grained details in close-up views and free of aliasing in distant views while maintaining efficient reconstruction, \ie within five minutes, and compact representation, \ie $25\%$ smaller model size than Instant-ngp.
This is realized by our novel $\trimip$ encoding and cone casting.
The $\trimip$ encoding featurizes the 3D space by three mipmaps to model the pre-filtered 3D feature space, such that the sample spheres from the cone casting can be encoded in an area-sampling manner.
We also develop a hybrid volume-surface rendering strategy to enable real-time rendering (> 60 FPS) on consumer-level devices.
Extensive quantitative and qualitative experiments demonstrate our Tri-MipRF achieves state-of-the-art rendering quality while having a super-fast reconstruction speed.
Also, the reconstruction results on the in-the-wild captures demonstrate the applicability of our Tri-MipRF.



\clearpage

{\small
	\bibliographystyle{ieee_fullname}
	\bibliography{egbib}
}

\clearpage
\appendix
\newcommand{\no}{{\color{red}\ding{55}}}
\newcommand{\yes}{{\textcolor[rgb]{0,0.75,0}{\checkmark}}}
\newcommand{\fair}{{\color{gray}{\checkmark\kern-1.1ex\raisebox{1.0ex}{\rotatebox[origin=c]{125}{--}}}}}

\onecolumn 
\begin{center}
\textbf{\Large Supplementary Material}
\vspace{1em}
\end{center}

\section{Characteristics Matrix}

We compared various crucial characteristics of cutting-edge NeRF methods and our Tri-MipRF in Tab.~\ref{table:feature_comp}, including quality aspect, \eg, anti-aliasing, and efficiency aspect, \eg, fast reconstruction, real-time rendering, and compact model. We can see that, except for our Tri-MipRF, none of them can support anti-aliasing, fast reconstruction, real-time rendering, and compact model, at the same time. These positive characteristics are enabled by our $\trimip$ encoding and cone casting.

\begin{table}[h]
	\centering
	\renewcommand{\tabcolsep}{10pt}
	\begin{tabular}{lcccc}
		\toprule
		Method                   & Anti-aliasing            &Fast Reconstruction & Real-time Rendering & Compact Model                              \\
		\midrule
		NeRF~\cite{mildenhall2020nerf}          & \no  & \no   & \no &  \yes 
		\\
		MipNeRF~\cite{barron2021mip}       & \yes   & \no   & \no &  \yes 
		\\
		Instant-ngp~\cite{muller2022instant}   & \no  & \yes   & \fair &  \yes 
		\\
        TensoRF~\cite{chen2022tensorf}       & \no  & \fair   & \fair &  \yes 
		\\
		PlenOctrees~\cite{yu2021plenoctrees}	  & \no  & \no   & \yes &  \no 
		\\
        Plenoxels~\cite{fridovich2022plenoxels}	  & \no  & \yes   & \fair &  \no 
		\\
		\emph{Tri-MipRF (ours)}   & \yes  & \yes   & \yes &  \yes 
		\\
		\bottomrule
	\end{tabular}
	\vspace{1em}
	\caption{Characteristics matrix of cutting-edge NeRF methods and ours. ``\yes'' means ``yes'', ``\fair'' means ``moderate'', and ``\no'' means ``no''.}
	\label{table:feature_comp}
\end{table}

\section{Model Details}

\subsection{Tiny MLP}
\label{subsec:mlp}
The goal of the tiny MLP is to nonlinearly map the feature vector $\feat$ produced by the $\trimip$ encoding and the view direction $\viewdir$ to density $\density$ and color $\col$ of the sampled sphere $\sphere$.
The feature vector $\feat$ has a dimension of $48$ since the mipmaps $\mipmap$ in $\trimip$ encoding has a shape of $512 \times 512\times 16$ as described in Sec.4.1 of the main paper. 
The first two layers of the MLP take $\feat$ as input and produce the density $\density$ and a geometric feature $\feat_\text{geo}$ with a dimension of $15$.
And the view direction $\viewdir$ is encoded by the spherical harmonics basis and then fed into the last three layers the MLP together with the $\feat_\text{geo}$ to estimate the final view-dependent color $\col$, which is similar to~\cite{muller2022instant}.
The width of the tiny MLP is empirically set to $128$.
The activation function of all the layers is the ReLU, except for the output layer of density $\density$, where we adopt the truncated exponential function followed~\cite{muller2022instant}.
This shallow MLP is implemented with tiny-cuda-nn that is well-optimized for fused and half-precision MLP.

\subsection{Optimization}
The optimizable parameters of our Tri-MipRF include the model weights $\modelweights$ of the tiny MLP and the mipmaps $\mipmap$ in the $\trimip$ encoding.
The model weights $\modelweights$ is initialized by the method proposed in~\cite{glorot2010understanding}, while the mipmaps $\mipmap$ is initialized by a uniform distribution of the interval $[-0.01, 0.01]$ to encourage the sparsity of $\mipmap$.   
We employ the AdamW optimizer~\cite{loshchilov2019decoupled} to train $\modelweights$ and $\mipmap$, where we set base learning rate for $\modelweights$ and scale up the base learning rate $10\times$ for $\mipmap$ since $\mipmap$ is a direct representation the reconstructed scene.
We set the base learning rate to $2 \times 10^{-3}$ and scale it up $0.6\times$ at steps $12K$, $18K$, $20K$, and $22K$, while the total number of iteration is $25K$.
And followed~\cite{muller2022instant}, we adopt the dynamic batch-size strategy that keep the total number of spheres in a batch to be roughly $256K$.
We will release our source code for better reproducibility upon publication.


\section{Detailed Results}

\begin{figure*}[t] 
	\centering
	\includegraphics[width=0.78\linewidth]{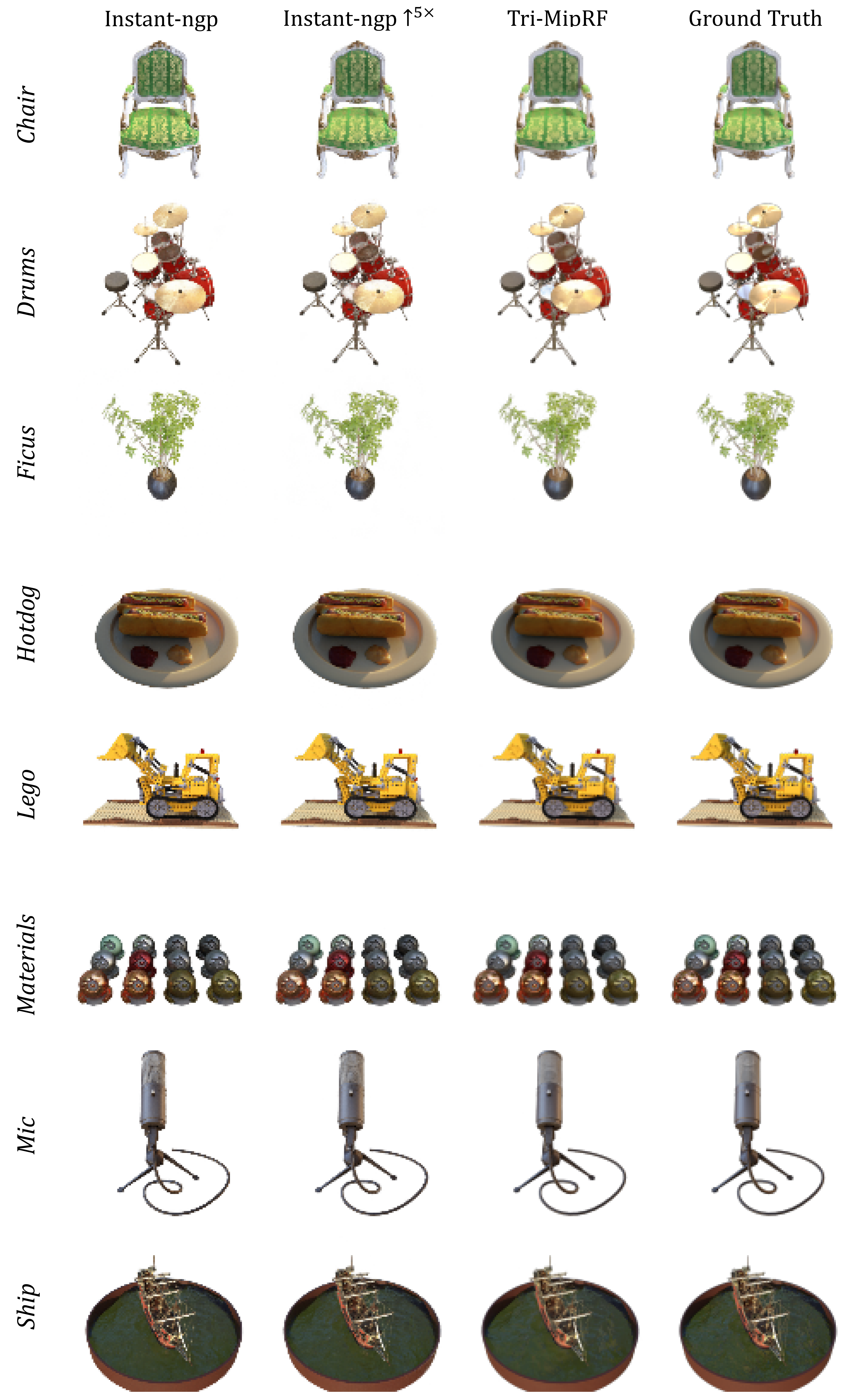}
	\caption{
		Qualitative $\nicefrac{1}{8}$ resolution rendering results of Instant-ngp~\cite{muller2022instant}, Instant-ngp $\uparrow^{5\times}$, and our Tri-MipRF on the multi-scale Blender dataset.
	}
		\vspace{-2em}
	\label{fig:qualitative_aliasing_supp}
\end{figure*}

\paragraph{Multi-scale Blender Dataset}
To demonstrate more detailed per-scene results of our Tri-MipRF, compared with other cutting-edge methods, we provide quantitative results of them under three metrics in Tab.~\ref{tab:avg_multiblender_perscene}.
We can see our Tri-MipRF outperforms all the other methods on almost all the scenes.
Visual comparisons of the renderings from Instant-ngp, Instant-ngp $\uparrow^{5\times}$, and our Tri-MipRF can be found in Fig.~\ref{fig:cont} and Fig.~\ref{fig:qualitative_aliasing_supp}. Our method consistently renders more fine-grained and anti-aliased images compared with Instant-ngp and Instant-ngp $\uparrow^{5\times}$. The aliasing artifacts are not easy to be observed in still pictures, so readers are highly recommended to watch the supplemental video for better evaluation.

\begin{table*}[h]
    \centering
    \small
    \begin{tabular}{l|cccccccc|c}
	 & \multicolumn{9}{c}{\textbf{PSNR}} \\
 & \scenename{chair}  & \scenename{drums}  & \scenename{ficus}  & \scenename{hotdog}  & \scenename{lego}  & \scenename{materials}  & \scenename{mic}  & \scenename{ship} & \scenename{Average} 
 \\ 
 \hline 
NeRF w/o $\mathcal{L}_\text{area}$&                    29.92  &                    23.27  &                    27.15  &                    32.00  &                    27.75  &                    26.30  &                    28.40  &                    26.46  & 27.66
\\
NeRF~\cite{mildenhall2020nerf} &                    33.39  &                    25.87  &                    30.37  &                    35.64  &                    31.65  &                    \cellcolor{yellow}30.18  &                    32.60  &                    30.09  & 31.23
\\
MipNeRF~\cite{barron2021mip} &                    \cellcolor{orange}37.14  &                    \cellcolor{orange}27.02  &                    \cellcolor{orange}33.19  &                    \cellcolor{orange}39.31  &                    \cellcolor{orange}35.74  &                    \cellcolor{red}32.56  &                    \cellcolor{orange}38.04  &                    \cellcolor{orange}33.08  & \cellcolor{orange}34.51
\\
\hline
Plenoxels~\cite{fridovich2022plenoxels} &                    32.79  &                    25.25  &                    30.28  &                    34.65  &                    31.26  &                    28.33  &                    31.53  &                    28.59 & 30.34 
\\
TensoRF~\cite{chen2022tensorf} &                    32.47  &                    25.37  &                    31.16  &                    34.96  &                    31.73  &                    28.53  &                    31.48  &                    29.08  & 30.60
\\
Instant-ngp~\cite{muller2022instant} &                    32.95  &                    26.43  &                    30.41  &                    35.87  &                    31.83  &                    29.31  &                    32.58  &                    30.23  & 31.20
\\
Instant-ngp $\uparrow^{5\times}$ &                    \cellcolor{yellow}34.15  &                    26.79  &                    \cellcolor{yellow}31.50  &                    \cellcolor{yellow}36.47  &                    \cellcolor{yellow}32.51  &                    29.49  &                    \cellcolor{yellow}33.81  &                    \cellcolor{yellow}30.78 & \cellcolor{yellow}31.94  
\\
\hline
Tri-MipRF w/o $\mipmap$ &                    33.09  &                    \cellcolor{yellow}26.85  &                    31.07  &                    36.08  &                    32.09  &                    29.85  &                    32.66  &                    30.17  & 31.48
\\
Tri-MipRF (Ours) &                    \cellcolor{red}37.72  &                    \cellcolor{red}28.55  &                    \cellcolor{red}33.77  &                    \cellcolor{red}39.96  &                    \cellcolor{red}36.51  &                    \cellcolor{orange}32.35  &                    \cellcolor{red}38.06  &                    \cellcolor{red}33.59 & \cellcolor{red}35.06  
\\

\multicolumn{9}{c}{} \\
 & \multicolumn{9}{c}{\textbf{SSIM}} \\
 & \scenename{chair}  & \scenename{drums}  & \scenename{ficus}  & \scenename{hotdog}  & \scenename{lego}  & \scenename{materials}  & \scenename{mic}  & \scenename{ship} & \scenename{Average} 
\\ 
\hline 
NeRF w/o $\mathcal{L}_\text{area}$&                    0.944  &                    0.891  &                    0.942  &                    0.959  &                    0.926  &                    0.934  &                    0.958  &                    0.861  &0.927
\\
NeRF~\cite{mildenhall2020nerf} &                    0.971  &                    0.932  &                    0.971  &                    0.979  &                    0.965  &                    \cellcolor{yellow}0.967  &                    0.980  &                    0.900 &0.958  
\\
MipNeRF~\cite{barron2021mip} &                    \cellcolor{orange}0.988  &                    \cellcolor{orange}0.945  &                    \cellcolor{orange}0.984  &                    \cellcolor{orange}0.988  &                    \cellcolor{orange}0.984  &                    \cellcolor{red}0.977  &                    \cellcolor{red}0.993  &                    \cellcolor{orange}0.922  & \cellcolor{orange}0.973
\\
\hline
Plenoxels~\cite{fridovich2022plenoxels} &                    0.968  &                    0.929  &                    0.972  &                    0.976  &                    0.964  &                    0.959  &                    0.979  &                    0.892 & 0.955  
\\
TensoRF~\cite{chen2022tensorf} &                    0.967  &                    0.930  &                    0.974  &                    0.977  &                    0.967  &                    0.957  &                    0.978  &                    0.895  & 0.956 
\\
Instant-ngp~\cite{muller2022instant} &                    0.971  &                    0.940  &                    0.973  &                    0.979  &                    0.966  &                    0.959  &                    0.981  &                    0.904  & 0.959
\\
Instant-ngp $\uparrow^{5\times}$ &                    \cellcolor{yellow}0.979  &                    \cellcolor{yellow}0.943  &                    \cellcolor{yellow}0.978  &                    \cellcolor{yellow}0.982  &                    \cellcolor{yellow}0.972  &                    0.959  &                    \cellcolor{yellow}0.985  &                    \cellcolor{yellow}0.909  & \cellcolor{yellow}0.963
\\
\hline
Tri-MipRF w/o $\mipmap$ &                    0.971  &                    0.941  &                    0.974  &                    0.980  &                    0.967  &                    0.960  &                    0.980  &                    0.901  & 0.959 
\\
Tri-MipRF (Ours) &                    \cellcolor{red}0.990  &                    \cellcolor{red}0.957  &                    \cellcolor{red}0.986  &                    \cellcolor{red}0.989  &                    \cellcolor{red}0.986  &                    \cellcolor{orange}0.972  &                    \cellcolor{orange}0.992  &                    \cellcolor{red}0.935 & \cellcolor{red}0.976  
\\
\multicolumn{9}{c}{} \\
 & \multicolumn{9}{c}{\textbf{LPIPS}} \\
 & \scenename{chair}  & \scenename{drums}  & \scenename{ficus}  & \scenename{hotdog}  & \scenename{lego}  & \scenename{materials}  & \scenename{mic}  & \scenename{ship} & \scenename{Average} 
\\ 
\hline 
NeRF w/o $\mathcal{L}_\text{area}$&                    0.035  &                    0.069  &                    0.032  &                    0.028  &                    0.041  &                    0.045  &                    0.031  &                    0.095  & 0.052
\\
NeRF~\cite{mildenhall2020nerf} &                    0.028  &                    \cellcolor{yellow}0.059  &                    0.026  &                    \cellcolor{yellow}0.024  &                    0.035  &                    \cellcolor{yellow}0.033  &                    0.025  &                    \cellcolor{yellow}0.085  & 0.044
\\
MipNeRF~\cite{barron2021mip} &                    \cellcolor{orange}0.011  &                    \cellcolor{orange}0.044  &                    \cellcolor{red}0.014  &                    \cellcolor{red}0.012  &                    \cellcolor{orange}0.013  &                    \cellcolor{red}0.019  &                    \cellcolor{red}0.007  &                    \cellcolor{red}0.062  & \cellcolor{orange}0.026
\\
\hline
Plenoxels~\cite{fridovich2022plenoxels} &                    0.040  &                    0.070  &                    0.032  &                    0.037  &                    0.038  &                    0.055  &                    0.036  &                    0.104  & 0.051
\\
TensoRF~\cite{chen2022tensorf} &                    0.042  &                    0.075  &                    0.032  &                    0.035  &                    0.036  &                    0.063  &                    0.040  &                    0.112  & 0.054
\\
Instant-ngp~\cite{muller2022instant} &                    0.035  &                    0.066  &                    0.029  &                    0.028  &                    0.040  &                    0.051  &                    0.032  &                    0.095  & 0.047
\\
Instant-ngp $\uparrow^{5\times}$ &                    \cellcolor{yellow}0.025  &                    \cellcolor{yellow}0.059  &                    \cellcolor{yellow}0.023  &                    0.025  &                    \cellcolor{yellow}0.031  &                    0.049  &                    \cellcolor{yellow}0.023  &                    0.089 & \cellcolor{yellow}0.041 
\\
\hline
Tri-MipRF w/o $\mipmap$ &                    0.036  &                    0.066  &                    0.030  &                    0.028  &                    0.039  &                    0.051  &                    0.032  &                    0.099   & 0.048
\\
Tri-MipRF (Ours) &                    \cellcolor{red}0.010  &                    \cellcolor{red}0.042  &                    \cellcolor{red}0.014  &                    \cellcolor{red}0.012  &                    \cellcolor{red}0.012  &                    \cellcolor{orange}0.029  &                    \cellcolor{orange}0.008  &                    \cellcolor{red}0.062  &  \cellcolor{red}0.024 
\\
\multicolumn{9}{c}{}

    \end{tabular}
    \caption{Quantitative per-scene results on the test set of the multi-scale Blender dataset. For each scene, we report the arithmetic mean of each metric averaged over the four scales used in the dataset. The best, second-best, and third-best results are marked in red, orange, and yellow, respectively.
    }
    \label{tab:avg_multiblender_perscene}
\end{table*}

\paragraph{Single-scale Blender Dataset}
To show more detailed per-scene results of our Tri-MipRF, compared with other cutting-edge methods, on the single-scale Blender dataset, we provide quantitative results of them under three metrics in Tab.~\ref{tab:avg_singleblender_perscene}.
Even the single-scale Blender dataset observes the scene at a roughly constant distant, where the point-sampling-based methods would not suffers from the scale issue, our Tri-MipRF still outperforms them in terms all the three metrics.
It demonstrates the high applicability of our method for reconstructing objects at a constant or varying observing distances.

\begin{table*}[h]
    \centering
    \small
    \begin{tabular}{l|cccccccc|c}
	 & \multicolumn{9}{c}{\textbf{PSNR}} \\
 & \scenename{chair}  & \scenename{drums}  & \scenename{ficus}  & \scenename{hotdog}  & \scenename{lego}  & \scenename{materials}  & \scenename{mic}  & \scenename{ship} & \scenename{Average} 
 \\ 
 \hline 
SRN~\cite{srn}&                    29.96  &                    17.18  &                    20.73  &                    26.81  &                    20.85  &                    18.09  &                    26.85  &                    20.60  &  22.26
\\
LLFF~\cite{mildenhall2019local}&                    28.72  &                    21.13  &                    21.79  &                    31.41  &                    24.54  &                    20.72  &                    27.48  &                    23.22  &  24.88
\\
Neural Volumes~\cite{neuralvolumes}&                    28.33  &                    22.58  &                    24.79  &                    30.71  &                    26.08  &                    24.22  &                    27.78  &                    23.93  &   26.05
\\
Plenoxels~\cite{fridovich2022plenoxels} &                             33.98  &                    25.35  &                    31.83  &                    36.43  &                    34.10  &                    29.14  &                    33.26  & 29.62 &  31.71
\\
NeRF~\cite{mildenhall2020nerf} &                    34.17  &                    25.08  &                    30.39  &                    36.82  &                    33.31  &                    30.03  &                    34.78  &                    29.30  &  31.74
\\
DVGO~\cite{sun2022direct} &                    34.09  &                    25.44  &                    32.78  &                    36.74  &                    34.64  &                    29.57  &                    33.20  &                    29.13  &  31.95
\\
MipNeRF~\cite{barron2021mip} &                    \cellcolor{yellow}35.14  &                    25.48  &                    33.29  &                    \cellcolor{orange}37.48  &                    35.70  &                    \cellcolor{orange}30.71  &                    \cellcolor{orange}36.51  &                    \cellcolor{yellow}30.41  & 33.09
\\
TensoRF~\cite{chen2022tensorf} &                    \cellcolor{orange}35.76  &                    \cellcolor{yellow}26.01  &                    \cellcolor{orange}33.99  &                    \cellcolor{yellow}37.41  &                    \cellcolor{red}36.46  &                    \cellcolor{yellow}30.12  &                    34.61  &                    \cellcolor{orange}30.77  & \cellcolor{yellow}33.14
\\
Instant-ngp~\cite{muller2022instant} &                    35.00  &                    \cellcolor{orange}26.02  &                    \cellcolor{yellow}33.51  &                    37.40  &                    \cellcolor{orange}36.39  &                    29.78  &                    \cellcolor{yellow}36.22  &                    \cellcolor{red}31.10  & \cellcolor{orange}33.18
\\
\hline
Tri-MipRF (Ours) &                    \cellcolor{red}36.10  &                    \cellcolor{red}26.59  &                    \cellcolor{red}34.51  &                    \cellcolor{red}38.54  &                    \cellcolor{yellow}36.15  &                    \cellcolor{red}30.73  &                    \cellcolor{red}37.75  &                    28.78 & \cellcolor{red}33.65  
\\

\multicolumn{9}{c}{} \\
 & \multicolumn{9}{c}{\textbf{SSIM}} \\
 & \scenename{chair}  & \scenename{drums}  & \scenename{ficus}  & \scenename{hotdog}  & \scenename{lego}  & \scenename{materials}  & \scenename{mic}  & \scenename{ship} & \scenename{Average} 

\\ 
\hline 
SRN~\cite{srn}&                    0.910  &                    0.766  &                    0.849  &                    0.923  &                    0.809  &                    0.808  &                    0.947  &                    0.757  & 0.846
\\
LLFF~\cite{mildenhall2019local}&                    0.948  &                    0.890  &                    0.896  &                    0.965  &                    0.911  &                    0.890  &                    0.964  &                    0.823 & 0.911  
\\
Neural Volumes~\cite{neuralvolumes}&                    0.916  &                    0.873  &                    0.910  &                    0.944  &                    0.880  &                    0.888  &                    0.946  &                    0.784  & 0.893
\\
Plenoxels~\cite{fridovich2022plenoxels} &                    0.977  &                    0.933  &                    0.976  &                    0.980  &                    0.976  &                    0.949  &                    0.985  &                    \cellcolor{yellow}0.890  & 0.958
\\
NeRF~\cite{mildenhall2020nerf} &                    0.975  &                    0.925  &                    0.967  &                    0.979  &                    0.968  &                    \cellcolor{red}0.953  &                    0.987  &                    0.869  & 0.953
\\
DVGO~\cite{sun2022direct} &                    0.977  &                    0.930  &                    0.978  &                    0.980  &                    0.976  &                    0.951  &                    0.983  &                    0.879  & 0.957
\\
MipNeRF~\cite{barron2021mip} &                    \cellcolor{yellow}0.981  &                    0.932  &                    0.980  &                    \cellcolor{orange}0.982  &                    0.978  &                    0.959  &                    \cellcolor{orange}0.991  &                    0.882  & 0.961
\\
TensoRF~\cite{chen2022tensorf} &                    \cellcolor{red}0.985  &                    \cellcolor{orange}0.937  &                    \cellcolor{orange}0.982  &                    \cellcolor{orange}0.982  &                    \cellcolor{red}0.983  &                    \cellcolor{yellow}0.952  &                    0.988  &                    \cellcolor{orange}0.895  & \cellcolor{red}0.963
\\
Instant-ngp~\cite{muller2022instant} &                    0.979  &                    \cellcolor{orange}0.937  &                    \cellcolor{yellow}0.981  &                    \cellcolor{orange}0.982  &                    \cellcolor{orange}0.982  &                    0.951  &                    \cellcolor{yellow}0.990  &                    \cellcolor{red}0.896  & \cellcolor{red}0.963
\\
\hline
Tri-MipRF (Ours) &                    \cellcolor{red}0.985  &                    \cellcolor{red}0.939  &                    \cellcolor{red}0.983  &                    \cellcolor{red}0.984  &                    \cellcolor{orange}0.982  &                    \cellcolor{red}0.953  &                    \cellcolor{red}0.992  &                    0.879  & \cellcolor{red}0.963
\\

\multicolumn{9}{c}{} \\
 & \multicolumn{9}{c}{\textbf{LPIPS}} \\
 & \scenename{chair}  & \scenename{drums}  & \scenename{ficus}  & \scenename{hotdog}  & \scenename{lego}  & \scenename{materials}  & \scenename{mic}  & \scenename{ship} & \scenename{Average} 
\\ 
\hline 
SRN~\cite{srn}&                    0.106  &                    0.267  &                    0.149  &                    0.100  &                    0.200  &                    0.174  &                    0.063  &                    0.299  & 0.170
\\
LLFF~\cite{mildenhall2019local}&                    0.064  &                    0.126  &                    0.130  &                    0.061  &                    0.110  &                    0.117  &                    0.084  &                    0.218 & 0.114
\\
Neural Volumes~\cite{neuralvolumes}&                    0.109  &                    0.214  &                    0.162  &                    0.109  &                    0.175  &                    0.130  &                    0.107  &                    0.276  & 0.160
\\
Plenoxels~\cite{fridovich2022plenoxels} &                    0.031  &                    \cellcolor{yellow}0.067  &                    0.026  &                    0.037  &                    0.028  &                    0.057  &                    0.015  &                    0.134  & 0.049
\\
NeRF~\cite{mildenhall2020nerf} &                    0.026  &                    0.071  &                    0.032  &                    0.030  &                    0.031  &                    \cellcolor{orange}0.047  &                    \cellcolor{yellow}0.012  &                    0.150  & 0.050
\\
DVGO~\cite{sun2022direct} &                    0.027  &                    0.077  &                    0.024  &                    0.034  &                    0.028  &                    0.058  &                    0.017  &                    0.161  & 0.053
\\
MipNeRF~\cite{barron2021mip} &                    \cellcolor{orange}0.021  &                    \cellcolor{red}0.065  &                    \cellcolor{red}0.020  &                    \cellcolor{orange}0.027  &                    0.021  &                    \cellcolor{red}0.040  &                    0.009  &                    \cellcolor{yellow}0.138  & \cellcolor{orange}0.043
\\
TensoRF~\cite{chen2022tensorf} &                    \cellcolor{yellow}0.022  &                    0.073  &                    \cellcolor{yellow}0.022  &                    0.032  &                    \cellcolor{yellow}0.018  &                    0.058  &                    0.015  &                    \cellcolor{yellow}0.138  & 0.047
\\
Instant-ngp~\cite{muller2022instant} &                    \cellcolor{yellow}0.022  &                    0.071  &                    0.023  &                    \cellcolor{orange}0.027  &                    \cellcolor{orange}0.017  &                    0.060  &                    \cellcolor{orange}0.010  &                    \cellcolor{red}0.132 & \cellcolor{yellow}0.045 
\\
\hline
Tri-MipRF (Ours) &                    \cellcolor{red}0.016  &                    \cellcolor{orange}0.066  &                    \cellcolor{red}0.020  &                    \cellcolor{red}0.021  &                    \cellcolor{red}0.016  &                    \cellcolor{yellow}0.052  &                    \cellcolor{red}0.008  &                    \cellcolor{orange}0.136  & \cellcolor{red}0.042
\\
\multicolumn{9}{c}{}

    \end{tabular}
	\caption{Quantitative per-scene results on the test set of the single-scale Blender dataset. The best, second-best, and third-best results are marked in red, orange, and yellow, respectively.
    }
    \label{tab:avg_singleblender_perscene}
\end{table*}

\begin{figure*}[h]
  \centering 
  \subfloat[][]{
  \includegraphics[width=0.97\linewidth]{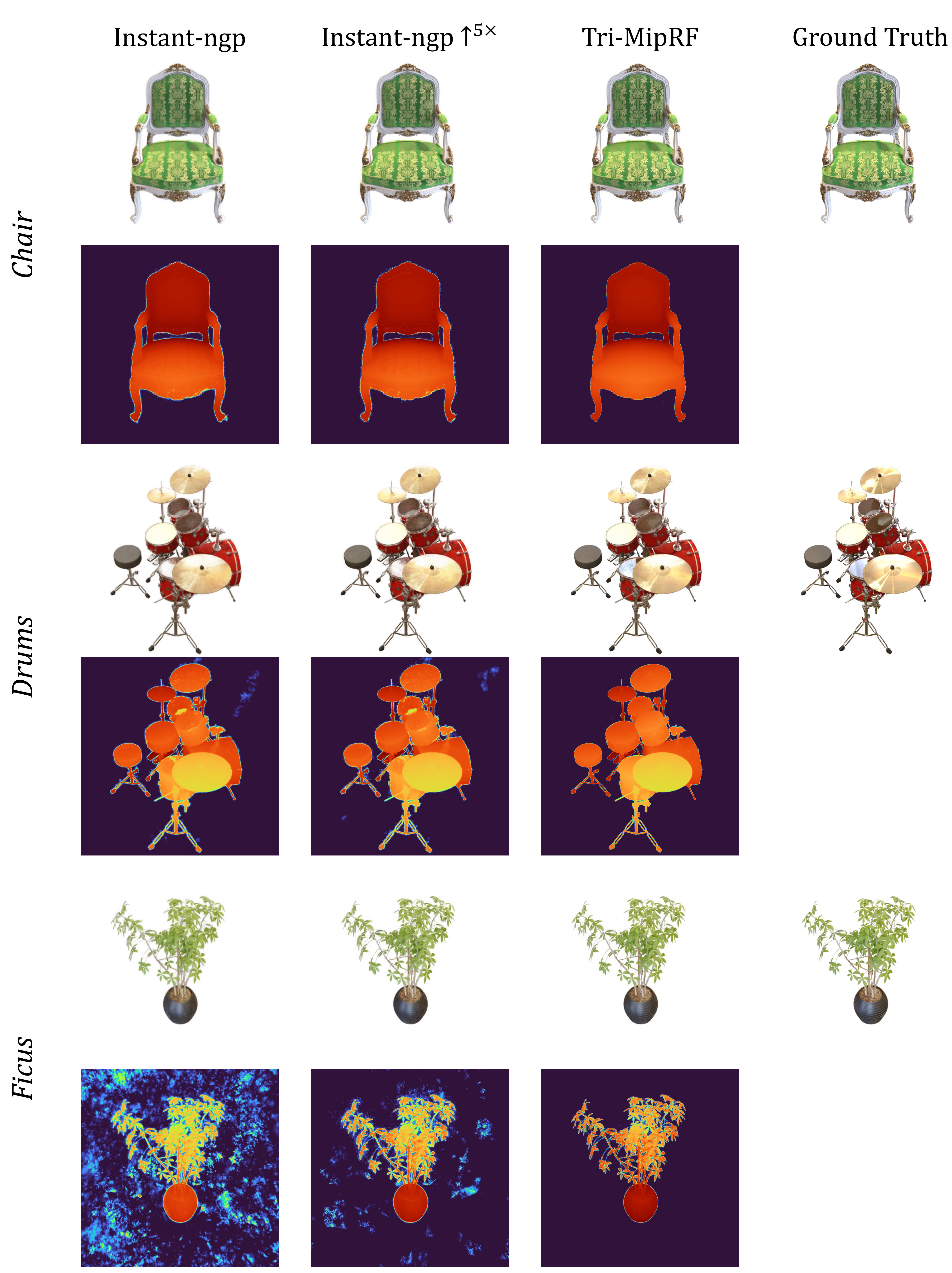}
  }%
  \label{fig:cont}
\end{figure*}

\begin{figure*}
  \ContinuedFloat 
  \centering 
  \subfloat[][]{
  \includegraphics[width=0.97\linewidth]{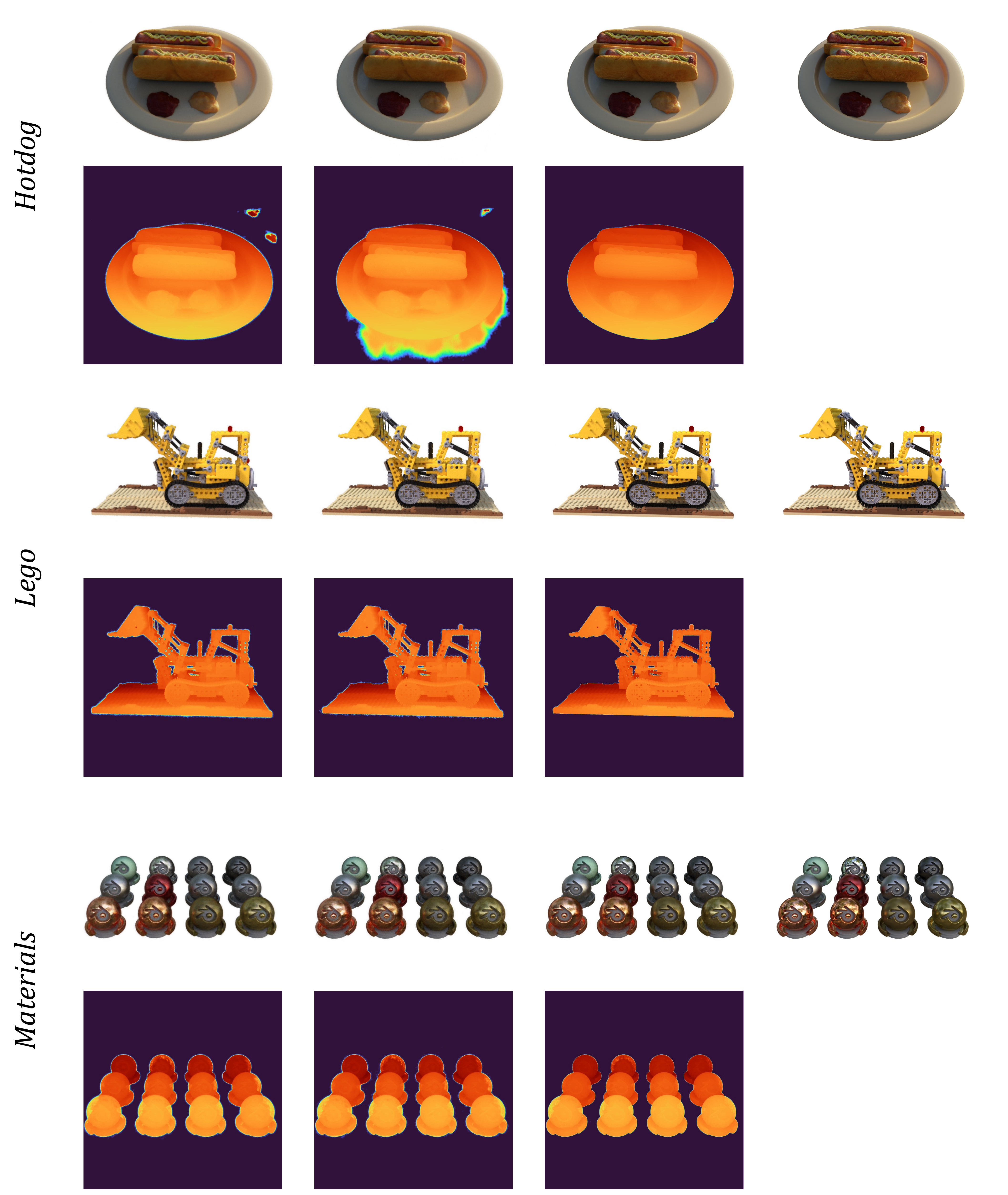}
  }%
  \vspace{2em}
  \label{fig:cont}
\end{figure*}

\begin{figure*}
  \ContinuedFloat 
  \centering 
  \subfloat[][]{
  \includegraphics[width=0.97\linewidth]{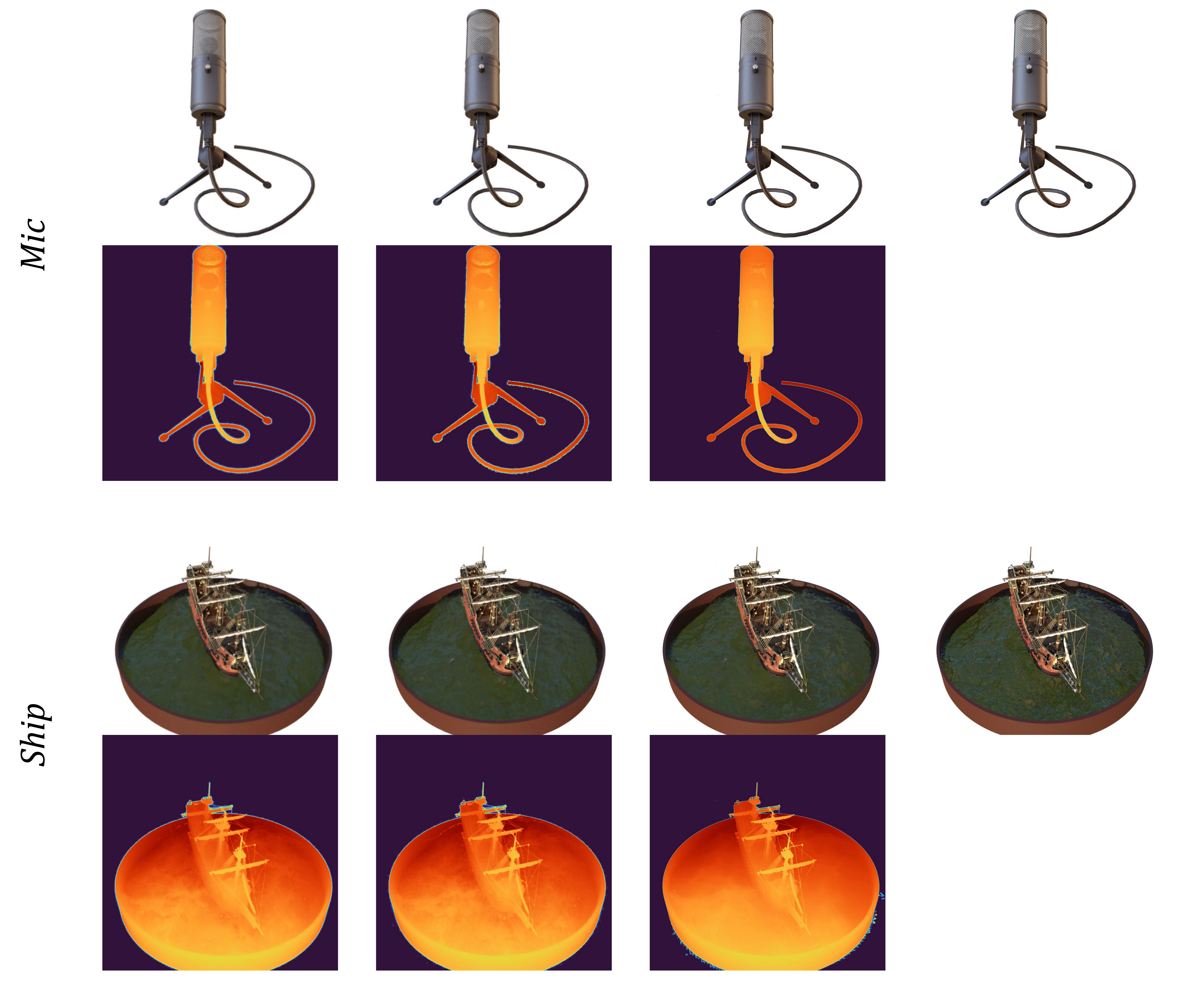}
  }%
  \vspace{1em}
  \caption[]{Qualitative full-resolution rendering results of Instant-ngp~\cite{muller2022instant}, Instant-ngp $\uparrow^{5\times}$, and our Tri-MipRF on the multi-scale Blender dataset. We show the rendered depth map under the RGB renderings.}
  \label{fig:cont}
\end{figure*}

\end{document}